\documentclass[letterpaper, 10 pt, conference]{ieeeconf}  

\IEEEoverridecommandlockouts                              

\usepackage[backend=biber,
            hyperref=true,
            url=false,
            isbn=false,
            doi=false,
            backref=false,
            style=ieee,
            citestyle=numeric-comp,
            sorting=none,%
            block=none]{biblatex}

\usepackage{amssymb}
\usepackage{amsmath}
\usepackage{amsfonts}
\usepackage{bbm}
\usepackage{booktabs}
\usepackage{dsfont}
\usepackage{multirow}
\usepackage{graphicx}
\usepackage{wrapfig}
\usepackage{siunitx}
\usepackage{subcaption}
\usepackage[table]{xcolor}
\usepackage{hyperref}

\usepackage[T1]{fontenc} 
\usepackage{setspace} 
\usepackage{dialogue}
\usepackage{calc} 
\usepackage{lipsum}

\newcommand\ours{MICoBot}
\newcommand{\rev}[1]{\textcolor{black}{#1}}

\newlength{\widestname}
\makeatletter
\renewenvironment{dialogue} {%
    \begin{list}{} {%
        \setlength\itemsep{\z@ \@plus .5ex}%
        \setlength{\parsep}{\parskip}%
        \setlength{\rightmargin}{0pt}
        \setlength{\labelwidth}{\widestname}
        \setlength{\labelsep}{0.5em}
        \setlength{\leftmargin}{\labelwidth}
        \addtolength{\leftmargin}{\labelsep}
      }%
      \PreDialogue\relax
    }{%
  \end{list}%
  }
\makeatother

\PassOptionsToPackage{compress}{natbib}

\newcommand{\speak}[2]{\item[\hfill {#1}] {\itshape ``{#2}''}}
\newcommand{\speaknoquote}[2]{\item[\hfill {#1}] {#2}}
\newcommand{\methodname}[0]{MICoBot}

\definecolor{LightBlue}{rgb}{0.88,0.88,1}
\definecolor{LightGreen}{rgb}{0.88,1,0.88}

\title{\LARGE \bf
Mixed-Initiative Dialog for Human-Robot Collaborative Manipulation
}

\author{
  Albert Yu$^{1, \dagger}$, Chengshu Li$^{2}$, Luca Macesanu$^{3}$, Arnav Balaji$^{1}$, Ruchira Ray$^{4}$, \protect\\ Raymond Mooney$^{1}$, Roberto Mart\'in-Mart\'in$^{1}$
\thanks{
    $^{1}$UT Austin, 
    $^{2}$OpenAI, 
    $^{3}$NYU, 
    $^{4}$University of Edinburgh,
    $^\dagger \texttt{\small albertyu@utexas.edu}$
  }
}

\begin{document}

\maketitle
\thispagestyle{empty}
\pagestyle{empty}

\begin{abstract}
Effective robotic systems for long-horizon human-robot collaboration must adapt to a wide range of human partners, whose physical behavior, willingness to assist, and understanding of the robot's capabilities may change over time. This demands a tightly coupled communication loop that grants both agents the flexibility to propose, accept, or decline requests as they coordinate toward completing the task effectively. 
We propose \textbf{\methodname{}}, a system that enables the human and robot, both using natural language, to take initiative in formulating, accepting, or rejecting proposals on who can best complete different steps of a task.
To handle diverse, task-directed dialog, and find successful collaborative strategies that minimize human effort, MICoBot makes decisions at three levels: (1) a meta-planner considers human dialog to formulate and code a high-level collaboration strategy, (2) a planner optimally allocates the remaining steps to either agent based on the robot's capabilities (measured by a simulation-pretrained affordance model) and the estimated human's willingness to help, and (3) an action executor decides the low-level actions to perform or words to say to the human.
In physical robot trials with 18 unique human participants, MICoBot significantly improves task success and user experience over a pure LLM baseline and standard agent allocation models.
See additional videos and materials at our project site.
\footnote{https://robin-lab.cs.utexas.edu/MicoBot/\label{proj-site}}.
\end{abstract}
\section{Introduction}
\label{sec:intro}

\begin{figure}
    \includegraphics[width=1\columnwidth]{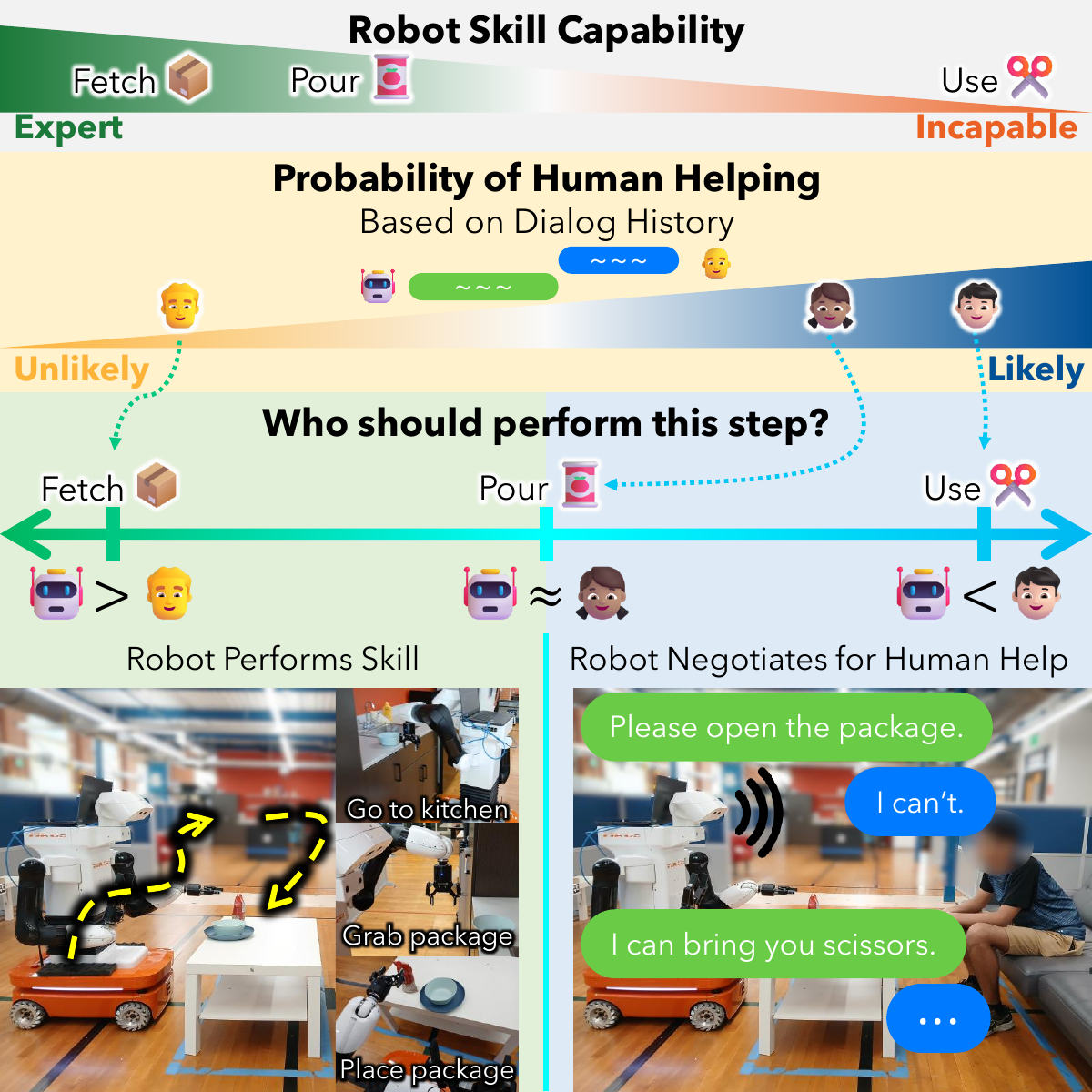}
    \caption{We present \ours{}, a system for human-robot collaboration where both agents can initiate and carry out physical and verbal actions. \ours{} uses both the robot's capability and the likelihood of human helping (inferred from previous dialog history) to determine whether the robot is better suited than the human to perform the skill. If it is, it attempts the skill itself. If not, it negotiates for human help.}
    \label{fig:pull-fig}
\end{figure}

Imagine preparing for a dinner party with a friend. Your friend might excel at mixing drinks while you focus on cooking the main dish. You are also better at decorating, while both of you reluctantly negotiate over less desirable tasks like cleaning.

Now, imagine a helper robot in place of the friend. Current robots are not fully autonomous for many household tasks, but they offer broad capabilities with varying levels of reliability that can be leveraged through collaboration with humans. To be an effective partner, such a robot must communicate in physically grounded natural language, decide when to take initiative or defer to the human, negotiate task allocation based on strengths and preferences, and adapt to changing contexts. These ingredients are essential not only for collaborative household robots, but also for coding assistants, chatbots, and AI agents more broadly.

Long-horizon tasks, such as preparing for a party, require dynamic, bidirectional collaboration across control, initiative, and communication. In particular, the ability to both take initiative and yield control is central to effective human–AI teamwork. However, current AI systems (e.g., chatbots) typically rely on one-directional, human-initiated interactions~\cite{ouyang2022training, achiam2023gpt}, while prior human–robot interaction (HRI) approaches often assume fixed collaboration plans and full human compliance~\cite{selvaggio2021autonomy}. Such assumptions limit flexibility and fail to account for the diverse preferences, capabilities, and strengths of different human partners. We argue that effective human–robot collaboration requires a paradigm shift toward mixed-initiative dialog as the communicative medium, enabling both agents to initiate, negotiate, and respond to proposals in natural language.


\begin{figure*}[htbp]
    \centering
    \includegraphics[width=1\textwidth]{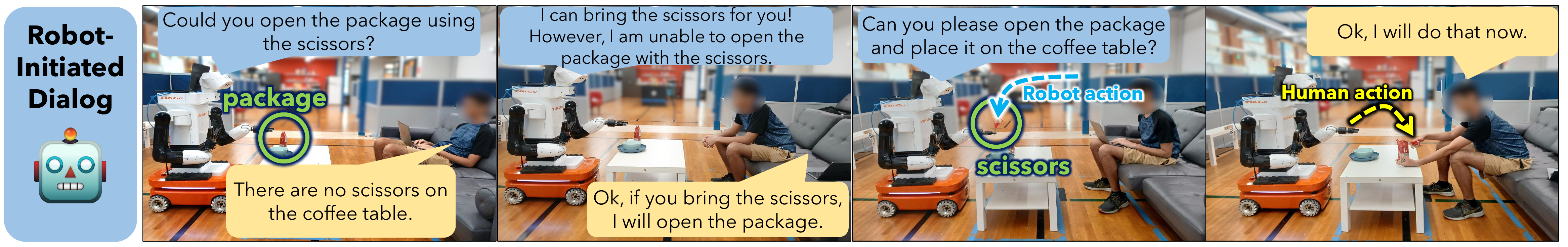}
    \includegraphics[width=1\textwidth]{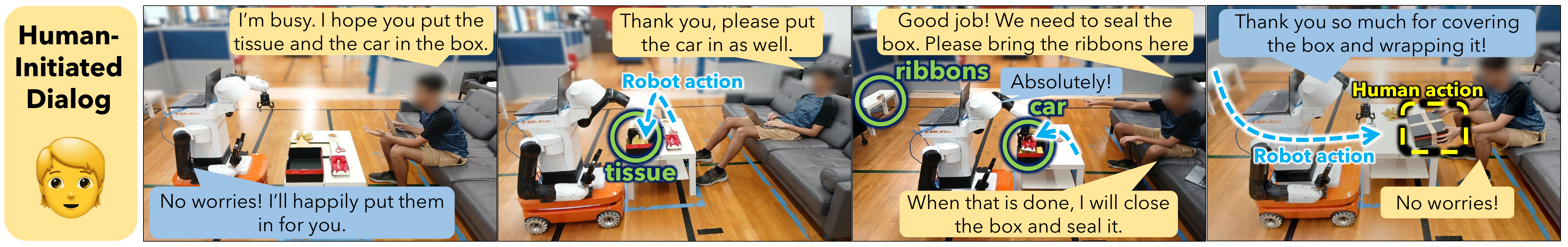}
    \caption{\ours{} supports \textit{both} robot-initiated (top row) \textit{and} human-initiated (bottom row) task-directed speech2speech dialog, where both agents discuss who is best suited to perform steps in a long-horizon task. These are real dialog and physical interactions from our user studies (see our
    website\textsuperscript{1}).}
    \label{fig:teaser}
    \vspace{-5mm}
\end{figure*}

To enable this paradigm shift, we introduce \ours{} (Mixed-Initiative Collaborative roBot), the first system that supports mixed-initiative dialog for seamless human–robot collaboration in the physical world. \ours{} allocates task steps to the most suitable agent (see Fig. ~\ref{fig:pull-fig}) in a way that maximizes overall success, minimizes human effort, and respects human-initiated requests. It achieves this by engaging in mixed-initiative dialog and negotiation to decide step allocation (see Fig.~\ref{fig:teaser}), while coordinating the physical and verbal actions required to execute the plan.


To realize this objective across diverse humans and long-horizon tasks, \ours{} optimizes decision-making at three levels. First, a meta-planner determines the high-level collaboration strategy, integrates human-specified preferences, and generates adaptive code for robot actions (verbal or physical). Second, a planner executes this code, selecting the optimal collaboration approach based on the environment state, a simulation-trained affordance model of robot capabilities, and a dynamic estimate of human helpfulness derived from prior interactions. Finally, an action executor carries out the next step of the plan, whether it involves manipulation, dialog initiation, or dialog response.


We validate \ours{} through extensive evaluation in both simulation and the real world. In simulation, we test against LLM-simulated humans with varying helpfulness and responsiveness; in real-world experiments, 18 participants collaborated with a TIAGo mobile manipulator on three household tasks. Our approach improves success rate by \textbf{50\%} compared to a pure LLM baseline and is preferred by over \textbf{75\%} of participants.


In summary, our contributions are:

\begin{itemize}

\item \textbf{A new problem setting} that integrates mixed-initiative natural language dialog with mixed-initiative human–robot interaction.

\item \textbf{A novel optimization framework} for task allocation that balances human and robot effort with success through a unified metric.

\item \textbf{A collaborative simulation environment} built on MiniBehavior~\cite{jin2023minibehavior}, featuring LLM-controlled virtual humans, with an interactive demo on our project site.\footnotemark[1]

\item \textbf{A hierarchical robotic system, \ours{}}, that enables mixed-initiative speech-to-speech human–robot collaboration and flexibly adapts to diverse real human collaborators in physically grounded, long-horizon tasks.

\end{itemize}
\section{Related Work}

\label{sec:related-work}

\textbf{Mixed-initiative dialog}~\cite{4081977, 796083, chu-carroll-2000-mimic} refers to communication with freeflowing questions and answers from both parties. In the NLP field, the dominant chatbot paradigm adopted by large language models (LLMs) largely eschews mixed-initiative interaction: humans pose substantive questions, and the chatbot primarily responds to fulfill these requests~\cite{ouyang2022training, achiam2023gpt}. 
Recent work has sought to make dialog systems more goal-directed and proactive by incorporating mixed-initiative strategies in tasks such as creating documents~\cite{collabllm2025}, persuading users to donate to charity, enhancing users’ emotional well-being~\cite{deng2023survey, yu2023promptbasedmontecarlotreesearch, chen2023controllable, deng2024plugandplaypolicyplannerlarge}, clarifying ambiguous human requests~\cite{qian2021databasesearchresultsdisambiguation, deng2023promptingevaluatinglargelanguage, chen2024learningclarifymultiturnconversations}, or as part of an active-learning framework~\cite{Thomason2019ImprovingGN}. 
However, none of these systems addressed mixed-initiative dialog in grounded, real-world collaborative scenarios involving physical manipulation tasks.



In the human-robot interaction (HRI) field, researchers have developed \textbf{human-robot collaboration systems} that interact through language but are restricted to \textbf{single-initiative dialog}.
Some of these systems integrate LLMs as task planners or delegators~\cite{wang2024mosaicmodularassistiveinteractive, mandi2023rocodialecticmultirobotcollaboration, feng2024largelanguagemodelbasedhumanagent} for tasks like real-world cooking~\cite{wang2024mosaicmodularassistiveinteractive} and object sorting~\cite{mandi2023rocodialecticmultirobotcollaboration}.
Other systems implement a leader-follower paradigm in simulated worlds, where the leader issues natural language instructions for the follower to execute~\cite{suhr2022executinginstructionssituatedcollaborative, kojima2021continuallearninggroundedinstruction, deepmindinteractiveagentsteam2022creatingmultimodalinteractiveagents, gao2023alexaarenausercentricinteractive}.
Single-initiative HRI systems can ask humans for clarification~\cite{ren2023robotsaskhelpuncertainty} or assistance~\cite{bennetot2020artificialagentsaskhelp,6630673, 10.5555/2832747.2832901}, or inform humans of their observations~\cite{chen:jair10, 4115652, cascianelli2018full}. 
However, by supporting only single-initiative dialog, these systems lack the capacity to adapt to the evolving nature of the human, robot, and environment—limiting their capacity to find the optimal division of labor that respects user preferences~\cite{mandi2023rocodialecticmultirobotcollaboration}. 

Some works in HRI have explored \textbf{mixed-initiative collaborative systems without dialog}, only with physical actions~\cite{4107804, natarajan2024mixedinitiativehumanrobotteamingsuboptimality, 9106686, rosero2021cooksunderstandingdynamichumanagent, paleja2024designsenablingcollaborationhumanmachine, 7379306}.
In particular, \cite{7451735} studied separate regimes of agent initiative (human-initiative, requesting help, or robot-initiative, proactively helping), but failed to support a natural human-robot dialog. 
By focusing solely on physical actions, these prior works overlook the critical role of communication in effective collaboration, thereby limiting the flexibility of the human-robot team. With \methodname{}, we enable both agents to take initiative—through both physical and verbal actions—via task-grounded dialog.


Several prior works in robotics and planning have studied the problem of \textbf{human-robot optimal task allocation}, typically optimizing the time to perform a task or minimizing idle agents, posing the problem as a scheduling optimization~\cite{vats2022synergisticschedulinglearningallocation, YU2021487}.
Others have prioritized different objectives, such as safety~\cite{Faccio2024} through the formulation of a constrained optimization problem~\cite{singh2023concurrentconstrainedoptimizationunknown}.
While these solutions may achieve shorter execution times, they assume a priori known capabilities and availability of all agents, including both robots and humans.
In contrast, \methodname{} can adapt to the specific human's willingness to help by estimating its availability based on previous dialog.


\section{Problem Setting: Task Collaboration with Mixed-Initiative Dialog}

\label{sec:prob-setting}

\vspace{-10pt}
\begin{figure}[ht!]
    \centering
    \includegraphics[width=\linewidth]{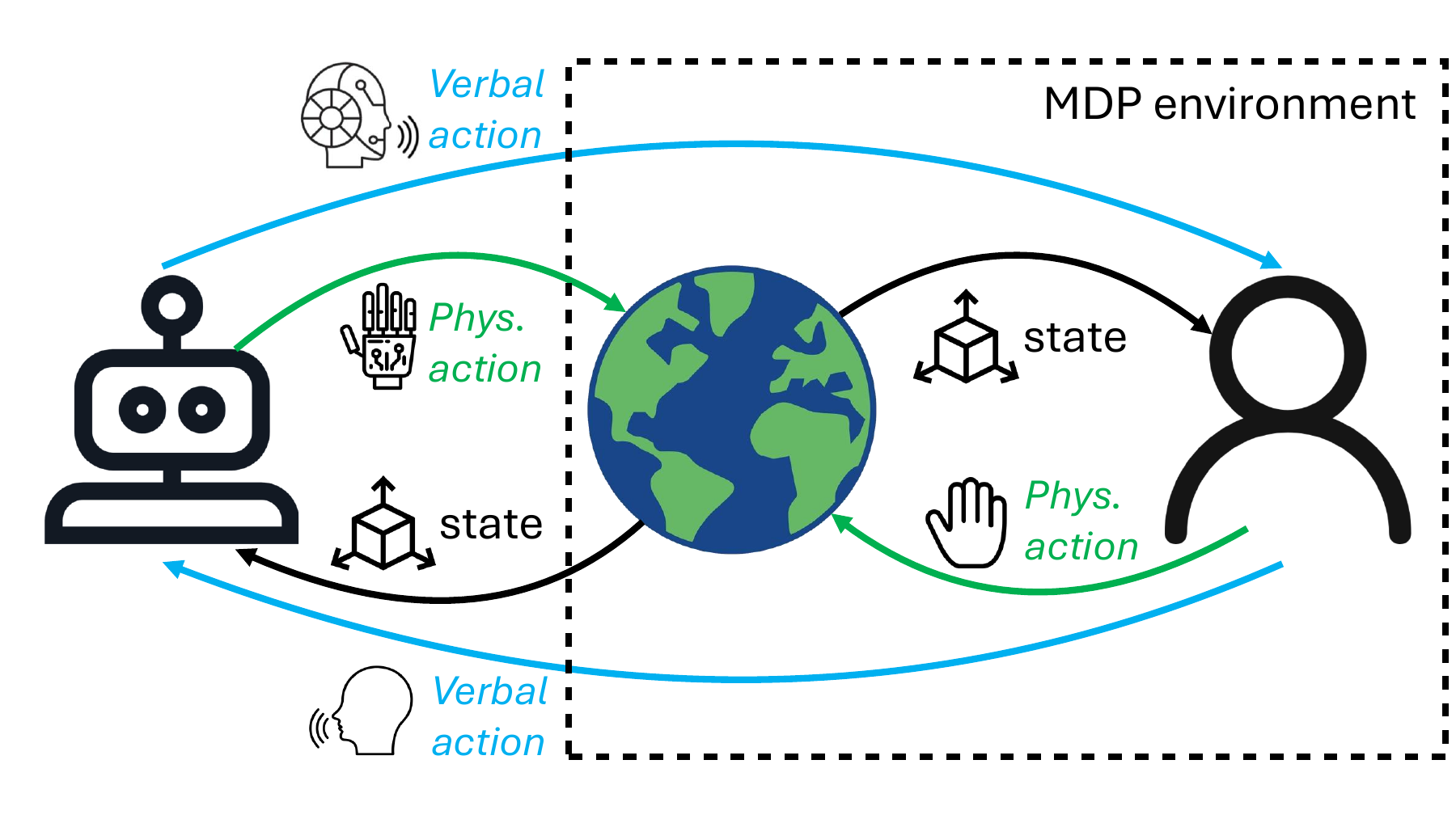}
    \caption{Proposed MDP for Mixed-Initiative Collaboration.}
    \label{fig:mdp}
\end{figure}

\textbf{MDP Formulation.} We study how human-robot collaborative manipulation can be facilitated through mixed-initiative dialog. We assume that both agents can observe the state of the world, $s\in\mathcal{S}$, and perform actions, 
$a\in\mathcal{A} = \mathcal{A}_p \cup \mathcal{A}_v$, comprised of a physical action space, $\mathcal{A}_p$ (e.g., move objects, open them, etc.), that directly affect the physical state of the environment $s$, and a free-form, natural language verbal action space, $\mathcal{A}_v$, which is directly observed by the other agent but does not change the physical state. 
We model the problem as a Markov Decision Process (MDP) from the robot's point of view (see Fig.~\ref{fig:mdp}), where on each environment step, the robot performs some action, $a_R \in \mathcal{A}_{p, R} \cup \mathcal{A}_{v, R}$ and receives an observation $o = [I, a_{v, H}, s_\mathit{proprio}]$ consisting of an RGB-D image $I$, an optional verbal action from the human partner $a_{v, H}$, and the robot's proprioceptive state $s_\mathit{proprio}$. 
Within each environment step, the human may perform a series of actions, $a_H \in \mathcal{A}_{p, H} \cup \mathcal{A}_{v, H}$, in its own physical and verbal action space after perceiving the world state and robot's previous dialog, $a_{v, R}$.

\textbf{Physical and Verbal Action Spaces.} The physical and verbal action spaces, $\mathcal{A}_p$ and $\mathcal{A}_v$, are shared between both agents. Each element of these action spaces is a parameterized action primitive represented by the pair, $a_{p/v}=(\omega_{p/v},\theta_{p/v})$. $\omega_p$ is the type of the physical action primitive (\texttt{open}, \texttt{pick-and-place}, etc.) and $\theta_p$ are the corresponding parameters (e.g., what object to open or pick and where to place it). 
We assume that humans are fully competent in executing all steps of a collaborative household manipulation task but may be unwilling or unavailable to perform some or all required actions.
Their behavior can range from indifferent (never acting) to overly proactive (completing the entire task without robot involvement).

In contrast, robots often have limited manipulation capabilities and may be unable to execute more complex actions, in which case it uses verbal actions to communicate with the human.
$\omega_v$ is the type of the verbal action primitive (\texttt{ask\_human\_for\_help}, \texttt{respond\_to\_human}, etc.), and $\theta_v$ are the corresponding parameters defining the context of the verbal primitive (e.g., what step the robot needs help on).
While the types of verbal actions are limited, each generates freeform and open-vocabulary language. 
\methodname{} first selects an abstract verbal action from this space, then translates it into a natural language utterance to negotiate with the human—conveying its requests and the assistance it requires for successful collaboration. This involves reasoning over asymmetric human and robot physical capabilities to devise collaboration strategies that maximize task success while minimizing human effort.

\textbf{Collaborative Task Definition and Problem Statement.} 
We assume the collaborative task is defined by a task plan of length $T$, known to both agents and represented as a sequence of unassigned \textbf{physical} action primitives, $[a_{p,0}, ..., a_{p, T-1}]$, such as [(\texttt{pick-and-place}(\texttt{box}, \texttt{table}), \ldots, \texttt{close}(\texttt{box})], obtained from the task instructions or off-the-shelf task planner. 
To complete the manipulation task while minimizing human effort, the system must allocate steps of the plan between the two agents—negotiating with the human through robot-initiated dialog to suggest assignments, adapting to human preferences through human-initiated dialog, and ultimately executing its assigned physical actions.
At each step $t$, the system must compute the best allocation of the remaining steps of the plan, $G = [g_t, ..., g_{T-1}]$, where $\forall t, g_t \in \{H, R\}$. 
The optimal allocation $G^*$ maximizes the expected task success probability while minimizing total human effort.
These objectives are inherently competing: a policy focused solely on maximizing success might allocate all steps to the human (assumed to be perfectly competent); conversely, minimizing human effort alone would assign all steps to the robot, even when it may be incapable of completing certain steps.
The optimization also incorporates constraints conveyed through the mixed-initiative dialog history, such as task allocation requests or proposed task splits. 
The resulting allocation $G^*$ determines whether the robot executes the current step ($R$) or negotiates with the human for assistance ($H$).

\section{\ours{}: Mixed-Initiative Collaborative Robot}
\label{sec:method}

\begin{figure*}[t!]
    \centering
    \includegraphics[width=1\textwidth,trim={0 0.0cm 0 0},clip]{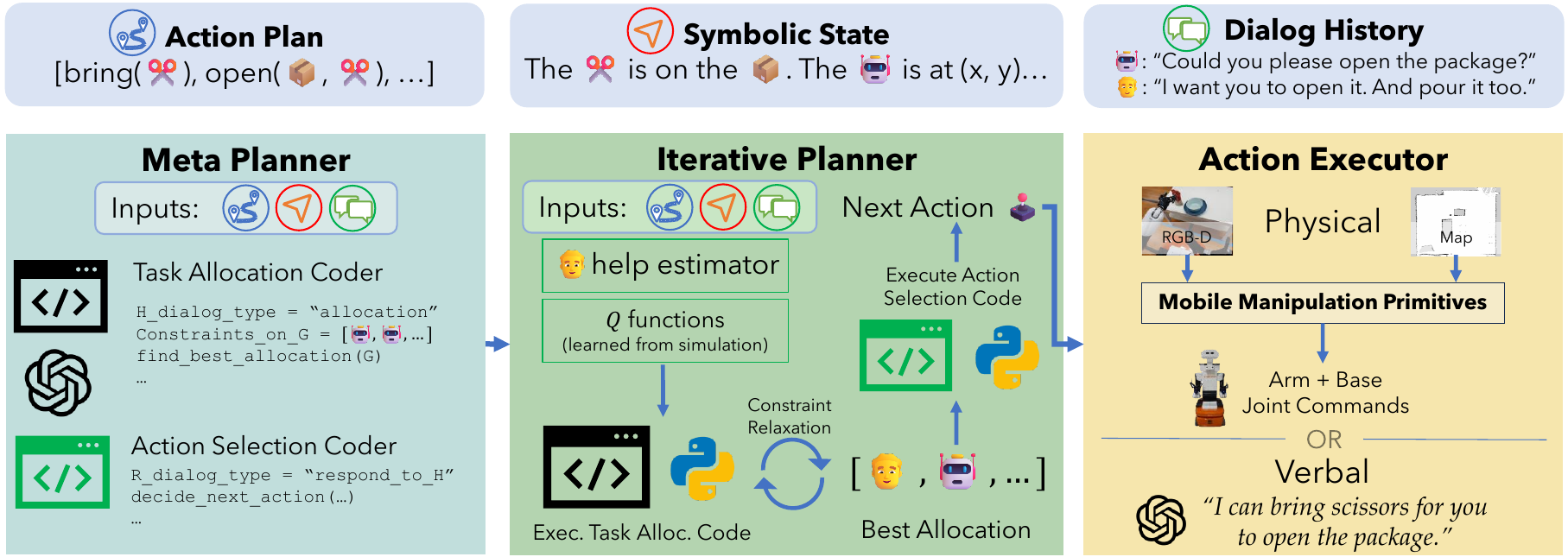}
    \caption{\ours{} consists of 3 decision-making modules: a meta-planner that produces a collaborative strategy expressed through adaptive planning code, a planner that executes the code and optimizes our objective (Eq.~\ref{eqn:objective}) to decide the next primitive action, and an action executor that outputs the low-level pose trajectory or verbal utterance to say to the human.
    }
    \label{fig:sys-arch}
\end{figure*}

\subsection{Collaborative Task Allocation as Optimization.} 

A helpful physical collaborator must aim for task success with minimal human effort while adhering to human preferences expressed in dialog (i.e., for certain steps to be done by a certain agent). Therefore, we formulate our objective for collaborative task allocation as a constrained optimization problem, where constraints are updated based on dialog exchanges.
To avoid a complex multi-objective formulation, we combine success probability and effort into a single Q-value by
building on prior work on temporal distances in RL~\cite{myers2025learningtemporaldistancescontrastive}. Then, to allocate task steps, \ours{} compares robot and human Q-values.

We assume each task step is executed by a multi-task policy $\pi$ that performs continuous low-level control at a fixed control frequency. In this low-level MDP (distinct from the high-level task MDP described in Sec.~\ref{sec:prob-setting}), we define the reward as $r = -1$ per time step until the skill completes or times out, at which point $r_{\mathit{termination}} = 0$. A well-trained Q-function, $Q: o_t \times a_t = (\omega_t, \theta_t) \mapsto \mathbb{R}$ with a discount factor of 1, thus represents the \textbf{negative expected number of timesteps} until skill completion from a given state.
For a perfectly competent agent (i.e., human), this corresponds to the average timesteps required to perform the action.
For an imperfect agent that may fail, the Q-function reflects a weighted expectation over both successful and failed outcomes—where failure contributes a significant timestep penalty (timeout) weighted by its probability. We assign each agent a distinct Q-function: $Q_R$ for the robot and $Q_H$ for the human.
These agent-specific Q-functions thus provide a unified, interpretable cost metric for comparing step allocations, jointly capturing both execution time (effort) and likelihood of success.


However, directly optimizing step allocation using only these two Q-functions diverges from realistic human-robot collaboration scenarios in three ways:
(1) human and robot effort are valued equally, ignoring the higher worth of human time and attention;
(2) the human is assumed to always comply with robot-initiated requests, overlooking variability in willingness or availability; and
(3) human-initiated requests or preferences are not considered, limiting the system’s adaptability to human intent.
To address (1), we introduce a \textit{human-effort factor}, $\alpha$, a ratio valuing human effort to robot effort. 
To address (2), human Q-values are adjusted with an inferred probability, $p_{H,t}$, of the human agreeing to perform action $a_{H,t} = \omega_t(\theta_t)$ when asked.
For less cooperative users, this probability lowers the expected success of $a_{H, t}$, effectively increasing the magnitude of the negative Q-value due to potential human refusal. To address (3), we enforce constraints, $C_1,\ldots,C_n$, extracted from human-initiated dialog—such as explicit requests to perform specific steps themselves or delegate them to the robot.
Altogether, we propose the following objective to find the optimal task allocation $G^*$:
\begin{equation}
\label{eqn:objective}
\begin{aligned}
\max_{g_t, ..., g_T} \quad & \sum_{t}^{T-1} \left( \mathds{1}_{g_t = H} \cdot \frac{\alpha}{p_{H, t}} + \mathds{1}_{g_t = R} \right) Q_{g_t}(s_t, a_t),\\
{s.t.} \quad & C_1, \ldots, C_n \text{ are satisfied}\\
\end{aligned}
\end{equation}

that minimizes expected time-to-success and human effort.

\subsection{\ours{} Framework}
\methodname{} is a three-level framework (Fig.~\ref{fig:sys-arch}) that includes 1) a meta-planner that processes human dialog and generates a collaborative strategy expressed in code, 2) an iterative planner that updates planning state variables and allocates and decides the next action to perform by executing the code, and 3) an action executor that carries out the action primitive, either through low-level physical actions or by formulating a dialog utterance to the human.

\textbf{L1: Meta-planner.} The meta-planner produces adaptive planning code that dictates the overall strategy for L2 and L3 to follow. 
Based on the most recent human dialog, the current symbolic state of the world, the task plan, and approximately 15 in-context learning (ICL) examples, it generates two pieces of code: first, \textbf{task allocation} code to adapt the optimization computation, such as to map human dialog into additional constraints, and second, \textbf{action selection} code for how to choose the next action, such as whether to engage in additional dialog, proposals to split up the task with the human, or negotiation rounds before proceeding further with physical steps in the plan. The meta-planner is implemented as an LLM-based (GPT-4o) coder, and prompts can be found on our project website.\footnotemark[1]

\textbf{L2: Iterative Planner.} 
The iterative planner runs code from the meta-planner (L1) to make two key decisions: whether to initiate dialog and which verbal or physical action to perform. L2 runs in two stages. \textit{First}, it performs constrained optimization to find the best task allocation for the remaining task steps. To do this, we evaluate Eq.~\ref{eqn:objective}  under all possible task allocations fulfilling the constraints. Initially, the planner attempts to incorporate all constraints from the mixed-initiative dialog history. If no feasible allocation is found (e.g., human asked the robot perform a step it is incapable of), the planner iteratively relaxes the most recent constraint, and the robot verbally explains its incapability.

To quantify Eq.~\ref{eqn:objective}, \methodname{} requires accurate Q-functions that capture each agent’s expected effort and success probability on each task step. To collect data to learn the robot's Q-function ($Q_R$), we use the OmniGibson simulator~\cite{li2022behavior}, configured with a coarse model of the real-world task, environment,
and action primitives, recording both completion times and success rate. 
We train a supervised network as $Q_R$ that predicts the expected timesteps for an action primitive $a$ to succeed from a given symbolic state $o$.
Conditioning $Q_R$ on symbolic states minimizes the sim-to-real gap when we deploy the function to our real-world setting. When estimating the human's Q-function ($Q_H$), we assume perfect competence (i.e., no execution failures).
Thus, we simply obtain time estimates for each step from an LLM predicting how long a human needs to execute action $a_t = \omega_t(\theta_t)$, plus a travel time estimate based on human-object distances.

To adapt to changing human helpfulness, \ours{} estimates the probability of human assistance at the current $t$-th timestep, $p_{H,t}$, using an LLM-based sentiment analysis over prior human-robot dialog. This enables \methodname{} to adapt to temporally-changing user sentiments. 
After deciding the optimal task allocation, the second stage of L2 executes meta-planner action selection code to generate the optimal action $a = \omega(\theta)$ to execute: a physical mobile manipulation primitive $\omega$ to perform a task step on objects specified in $\theta$, or a verbal primitive $\omega$ to initiate dialog to ask for help, propose splitting up steps, or respond to human-initiated dialog regarding specific task steps specified in $\theta$.

\textbf{L3: Action Executor.} The action executor performs the action primitive selected by the planner (L2). For \textit{physical actions}, it generates a trajectory for navigation and arm movement to reach the location of and manipulate the target object while avoiding obstacles. We build a pipeline similar to \cite{shah2024bumble} that uses the \texttt{move\_base} ROS package for navigation path planning over a 2D occupancy map, and Grounding DINO~\cite{liu2023grounding} to segment the target object from an open-world scene based on the object specified in $\theta$. We backproject segmented image pixels from RGB-D camera data into a 3D point cloud to identify graspable or placeable points in the robot’s workspace that the arm reaches through inverse kinematics (IK). For \textit{verbal actions}, an LLM generates free-form natural language utterances to communicate with the human based on the dialog intent $\omega$ and verbal action parameters $\theta$ decided in L2. The LLM uses 10 ICL examples to generate language grounded in the context of the task and dialog.

\textbf{Hierarchical Plan.} To streamline communication for long-horizon task plans, \methodname{} groups adjacent low-level steps into semantically meaningful abstract actions that can be discussed more succinctly with the human.
The system only descends to a finer-grained level of detail during negotiation over low-level step assignments, which reduces the frequency and complexity of dialog, resulting in more efficient and user-friendly communication.

\begin{table*}[t!]
\centering
\resizebox{\textwidth}{!}{
\begin{tabular}{lcccccccc}
\toprule
\multirow{2}{*}{} & 
\multicolumn{2}{c}{ \begin{tabular}{@{}c@{}}
    \textbf{Pour Package in Bowl} \\
    $n=6$
  \end{tabular}} & 
\multicolumn{2}{c}{ \begin{tabular}{@{}c@{}}
    \textbf{Assemble Toy Car} \\
    $n=6$
  \end{tabular}} & 
\multicolumn{2}{c}{ \begin{tabular}{@{}c@{}}
    \textbf{Pack Gift Box} \\
    $n=6$
  \end{tabular}} & 
\multicolumn{2}{c}{ \begin{tabular}{@{}c@{}}
    \textbf{Average} \\
    $n=18$
  \end{tabular}} \\
\cmidrule(lr){2-3} \cmidrule(lr){4-5} \cmidrule(lr){6-7} \cmidrule(lr){8-9}
 & \centering \ours{} & LLM & \ours{} & LLM & \ours{} & LLM & \ours{} (ours) & LLM \\
\midrule
Entire Task Success Rate (\%, $\uparrow$) & $\mathbf{100}$ & $83$ & $\mathbf{67}$ & $0$ & $\mathbf{67}$ & $0$ & $\mathbf{77.8 \pm 15.7}$ & $27.8 \pm 39.3$ \\
\% of task steps completed ($\uparrow$) & $\mathbf{100}$ & $93$ & $\mathbf{94}$ & $31$ & $\mathbf{88}$ & $50$ & $\mathbf{93.8 \pm 5.1}$ & $58.2 \pm 26.0$ \\
\% of steps performed by Human & $27$ & $29$ & $60$ & $5$ & $35$ & $21$ & $40.5 \pm 14.2$ & $18.2 \pm 9.7$ \\
\midrule
\% Users Preferring ... ($\uparrow$) & $\mathbf{67}$ & $33$ & $\mathbf{100}$ & $0$ & $\mathbf{67}$ & $33$ & $\mathbf{77.8}$ & $22.2$ \\
Communicative ability ($\uparrow, /5$) & $3.7$ & $\mathbf{3.8}$ & $\mathbf{4.3}$ & $1.3$ & $\mathbf{2.8}$ & $2.3$ & $\mathbf{3.6 \pm 1.0}$ & $2.5 \pm 1.4$ \\
\begin{tabular}[c]{@{}l@{}}Awareness of its Limitations\end{tabular} ($\uparrow, /5$) & $\mathbf{3.3}$ & $\mathbf{3.3}$ & $\mathbf{3.7}$ & $1.2$ & $\mathbf{4.2}$ & $2.5$ & $\mathbf{3.7 \pm 1.4}$ & $2.3 \pm 1.6$ \\
Overall Satisfaction working w/ Robot ($\uparrow, /5$) & $\mathbf{3.8}$ & $3.7$ & $\mathbf{3.5}$ & $1.5$ & $\mathbf{3.5}$ & $2.5$ & $\mathbf{3.6 \pm 0.8}$ & $2.6 \pm 1.4$ \\
\bottomrule
\end{tabular}
}
\caption{
Comparison between \ours{} (ours) and the LLM baseline across three real-world tasks on both objective (top 3 rows) and subjective (bottom 4 rows) metrics. Ratings out of 5 are on the Likert scale. Through more effective task allocation and communication, \ours{} achieves much higher task success rates and overall user satisfaction.}
\label{tab:real-world}
\end{table*}

\section{Experimental Evaluation}
\label{sec:exp}
We evaluate \methodname{} in the real-world, on a Tiago mobile manipulator working with 18 unique human participants on household tasks, and in simulation, on a collaborative framework we developed atop Mini-Behavior gridworld~\cite{jin2023minibehavior}. In our simulation framework, a robotic agent collaborates with a simulated human with parametrizable helpfulness and mood-varying dialog, which allows for larger-scale experimentation and controlled comparisons across methods across a wider range of human behavior and dialog dynamics. A successful robotic collaborator must achieve task success (our primary evaluation metric) while minimizing human effort (our secondary metric).
We also report \textbf{subjective measures of robot behavior}, including user satisfaction, preference rankings, and Likert-scale ratings.

\textbf{Environment.} In the real-world, we perform our experiments in a mock apartment with a kitchen and living room area with commonplace furniture. 
In all of our tasks, the robot and human work together on opposite sides of a coffee table. Simulating a household setting, the participant spends nearly all of their time on the couch, where they can do their personal (i.e., non-task-related) work. The human can be as inactive or proactive as they wish in performing physical and verbal actions as defined in Section~\ref{sec:prob-setting} (though we continue the trial if they initiate dialog beyond the scope). Each human user study consists of two 20-30 minute trials, in which they collaborate with both our method and a pure LLM baseline, ordered randomly. All trials \textbf{terminate} under any of the following conditions: a primitive fails irrecoverably, $4T$ 
steps have elapsed for a plan of length $T$, an infeasible step is allocated to the robot twice consecutively, or the human refuses twice to perform a step infeasible for the robot.

\textbf{Skills.} To perform long-horizon household tasks, the robot has access to several mobile-manipulation action primitives. \texttt{pick\_place\_mobile(obj, place\_loc)} moves to \texttt{obj} and places it atop \texttt{place\_loc}, another object in a potentially different room. \texttt{pour(obj, cont)} travels to \texttt{obj} and pours its contents into \texttt{cont}. Finally, \texttt{fold(obj)} folds down box flaps.

To initiate and respond within mixed-initiative dialog, the robot uses the following open-vocabulary verbal action primitives for dynamic collaboration with the human: \texttt{ask\_for\_human\_help} on a \texttt{step} the human is best suited to perform, \texttt{propose\_split} to split \texttt{steps} with the human, \texttt{explain\_incapability} to ask the human to perform a \texttt{step} that the robot can't perform, and \texttt{respond\_to\_human} to \texttt{accept/reject} requests the robot is capable/incapable of executing.

\textbf{Baselines.} Because multiple components of our method are powered by LLMs, we compare our approach to a pure LLM baseline (\textbf{LLM}) given the same information as our meta-planner: symbolic state, dialog history, task plan, and $\alpha$ human-robot effort tradeoff factor. The LLM baseline is also provided with a list of the robot’s available skills and assumes that the human always successfully completes a step once they agree to perform it. The LLM baseline is prompted to produce a plan allocation $G$ that primarily optimizes for task success and secondarily minimizes human effort.

To control for the amount of human effort in the user studies, we compute an additional random allocation baseline that does not involve a human participant, \textbf{RECB} (random effort-controlled baseline). Let $p_c$ denote the proportion of steps done by the human in our method's trials. RECB randomly allocates the current step to the human with probability $p_{c}$ and assumes a perfectly helpful human and oracle robot primitives with 100\% success rate.

In simulation, we additionally compare against an \textbf{RL} baseline (hierarchical task allocator + robot policy)
and a naive \textbf{Random} baseline allocating either agent (with probability 50\%) to perform the next step.


\textbf{Ablations.} To measure the importance of mixed-initiative dialog, we perform the following ablations in simulation: \textbf{H-init} and \textbf{R-init}, where the human or the robot alone, respectively, can initiate any dialog. We further ablate components of \ours{} in simulation by running it \textbf{w/o P\_H} (no $p_{H, t}$ estimation) and \textbf{w/o Plan Hierarchy} (where our method talks to the human in terms of granular, low-level steps instead of more understandable, high-level subtasks).


\textbf{Tasks.} We perform user studies on 3 real-world tasks, each with 6 users for a total of 18 unique participants.
\textbf{(1) Pour package into bowl}: bring the bowl, package, and scissors from the kitchen, cut open the package, and pour it into the bowl.
\textbf{(2) Assemble toy car}: bring the car parts, wheels, and drill from the shelf to the coffee table, drill in the wheels, switch the drill bit, and finally drill in the windows and seats.
\textbf{(3) Pack gift box}: fold the gift box, put tissue wrapping paper and a toy car in the box, close the lid, wrap ribbons, and tape down a gift bow.
Each task is 5 to 8 mobile manipulation steps long and requires varying degrees of human involvement.





\begin{figure}[t!]
\centering
\includegraphics[width=\linewidth]{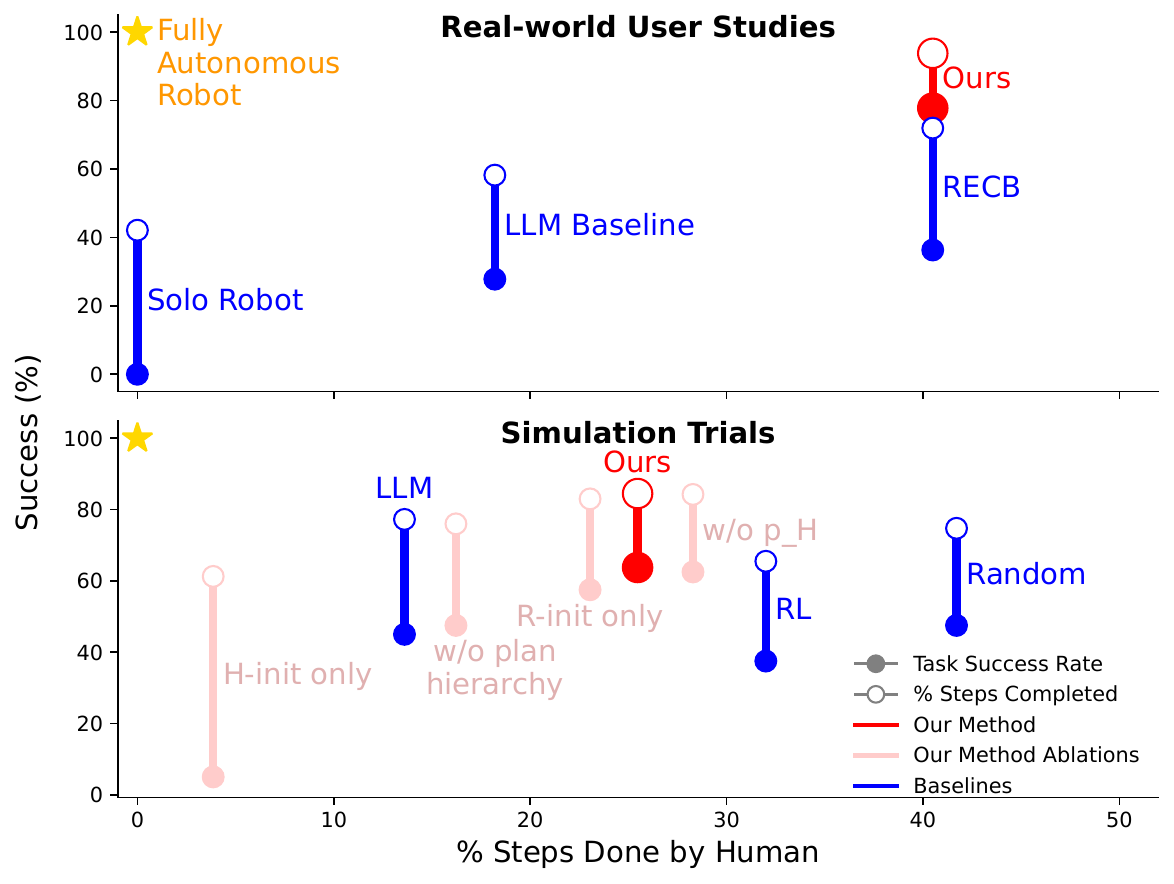}
\caption{In both \textbf{real-world} user studies \textbf{(top)} and \textbf{simulation trials} with a simulated human \textbf{(bottom)}, our method (red) demonstrates the best tradeoff in achieving task success (y-axis) for a given amount of human effort (x-axis) than baselines (blue) and our method's ablations (pink).}
\vspace{-3mm}
\label{fig:rw-success-vs-human-effort}
\end{figure}

\textbf{Experimental analysis.} Our experiments are designed to answer the following research questions:

\textbf{(1) Does our method achieve the best trade-off between task success and minimizing human effort?}
In our real-world user study (Table~\ref{tab:real-world}), \methodname{} achieves a 78\% task success rate compared to 28\% for the LLM baseline (statistically significant with $p$-value 0.007 under Fisher's exact test). Additionally, \methodname{} achieves a 94\% task step completion rate compared to the baseline's 58\% (statistically significant with $p$-value 0.002 under the Wilcoxon-signed-rank test). \ours{} understood its own limitations (through affordance functions trained in simulation), and was hence better at leveraging human assistance effectively on the steps it was ill-suited to perform.
The LLM baseline tended to prioritize minimizing human effort over task completion by allocating the robot multiple steps it was incapable of, since the LLM lacked an understanding of the robot's affordances.
Our method elicited more human effort than the baseline (40\% vs 18\%), so to control for the amount of human effort received, we compare our method to RECB in Figure~\ref{fig:rw-success-vs-human-effort}.
Despite RECB assuming oracle robot primitives and a perfectly cooperative human, our method still significantly outperforms it by more effectively balancing between success and human workload.




\begin{figure}[t!]
    \centering
    \includegraphics[width=\linewidth]{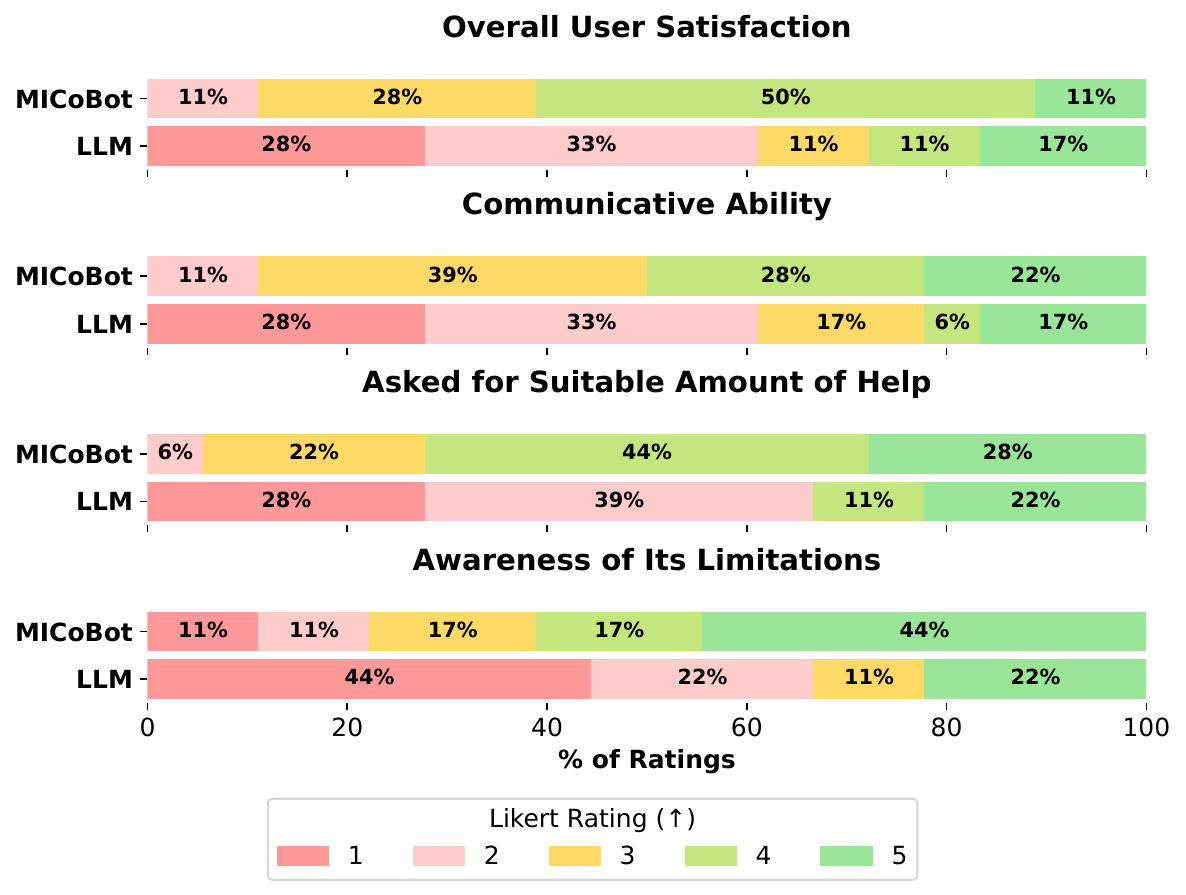}
    \caption{\rev{Our method substantially outperforms the pure LLM baseline in user ratings averaged over all $n=18$ participants.}}
    \label{fig:likert-ratings}
\end{figure}

\textbf{(2) How do users feel about working with our system?
} The A/B blind preference test in Table~\ref{tab:real-world} shows that 78\% of users preferred our method over the LLM baseline.
Our method also significantly outperformed the baseline in user scores on overall satisfaction, \rev{communicative ability, and capability in asking for a suitable amount of help (statistically significant under the Wilcoxon-signed-rank test with $p$-values ranging from 0.007 to 0.024; see Figure~\ref{fig:likert-ratings}).} In contrast, the LLM baseline often failed to express when it needed help and was unwilling to reject human requests it could not fulfill, leading to over-promises and task failures. A representative dialog exchange, available
on our project site,\footnotemark[1]
shows \methodname{} successfully persuading an initially reluctant user to perform a step the robot was incapable of executing.

\textbf{(3) Is mixed-initiative dialog critical to our method's performance?} Figure~\ref{fig:rw-success-vs-human-effort} (bottom) shows that our full method outperforms both ablated variants that restrict dialog to single-initiative modes: robot-only initiation (R-init) and human-only initiation (H-init). H-init performs especially poorly, as it prevents the robot from requesting help for steps it cannot execute. R-init performs slightly worse than the full method because it does not allow the human to proactively initiate dialog and assist when appropriate.
These results underscore the importance of mixed-initiative dialog in enabling flexible, robust human-robot collaboration.


In real-world user studies, \ours{} engaged in 2.4 dialog initiative shifts per trial, compared to the LLM baseline's 1.1. This enabled \ours{} to boost human acceptance of help requests from $55\%$ to $86\%$. The LLM baseline made far fewer help requests per trial (0.9 vs. \ours{}'s 2.9) and achieved a smaller acceptance increase ($70\%$ to $75\%$). This demonstrates mixed-initiative dialog is critical to collaborative discussion and task success.
\section{Conclusion}
\label{sec:conclusion}

We proposed \ours{}, a real-world robotic system that improves collaboration on long-horizon mobile manipulation tasks through mixed-initiative dialog with humans.
Our work unifies two previously unconnected lines of research: mixed-initiative dialog and HRI.
To this end, we formulated a novel optimization function and robotic framework using mixed-initiative dialog as a rich interface for task allocation to maximize task success while minimizing human effort and complying with verbally-expressed human preferences.
Real-world user studies with 18 human participants and extensive trials in simulation demonstrate the efficacy, adaptability, and user satisfaction of our method across a diverse range of human physical and verbal behavior.
\section{Limitations and Future Work}
\label{sec:limitations-future-work}
\ours{} represents our pioneering effort on facilitating mixed-initiative human-robot interaction through mixed-initiative natural-language dialog. While we focused on delegating steps for long-horizon manipulation tasks in a manner that maximizes task success and minimizes human effort, we believe this paper opens up exciting new avenues for future work. These include enabling both agents to learn to provide and incorporate spatial-temporal feedback to each other while performing a task, share relevant task information in an imperfect-information setting, and replan and redefine a task as necessary, all through mixed-initiative dialog.

\ours{} has a number of limitations.
First, it assumes that the human and robot work sequentially, and cannot handle cases where a robot and human wish to collaborate simultaneously on the same step in the plan, such as if the robot holds a roll of tape and the human cuts from it. 
Second, \ours{} assumes a fixed plan with a predetermined ordering of steps, where the human has a general, high-level understanding of the plan (but not the low-level steps that the robot plans over).
Our method could be improved with a more nuanced definition of ``effort'' beyond our time-based metric.
Finally, $p_{H, t}$ prediction can be improved, such as by processing tone-of-voice and facial expressions, to enable producing more emotionally understanding dialog, which can improve task success and user satisfaction.


\appendix
\renewcommand{\thesection}{Appendix \Alph{section}}

\begin{figure}[t!]
\centering
\includegraphics[width=0.32\linewidth]{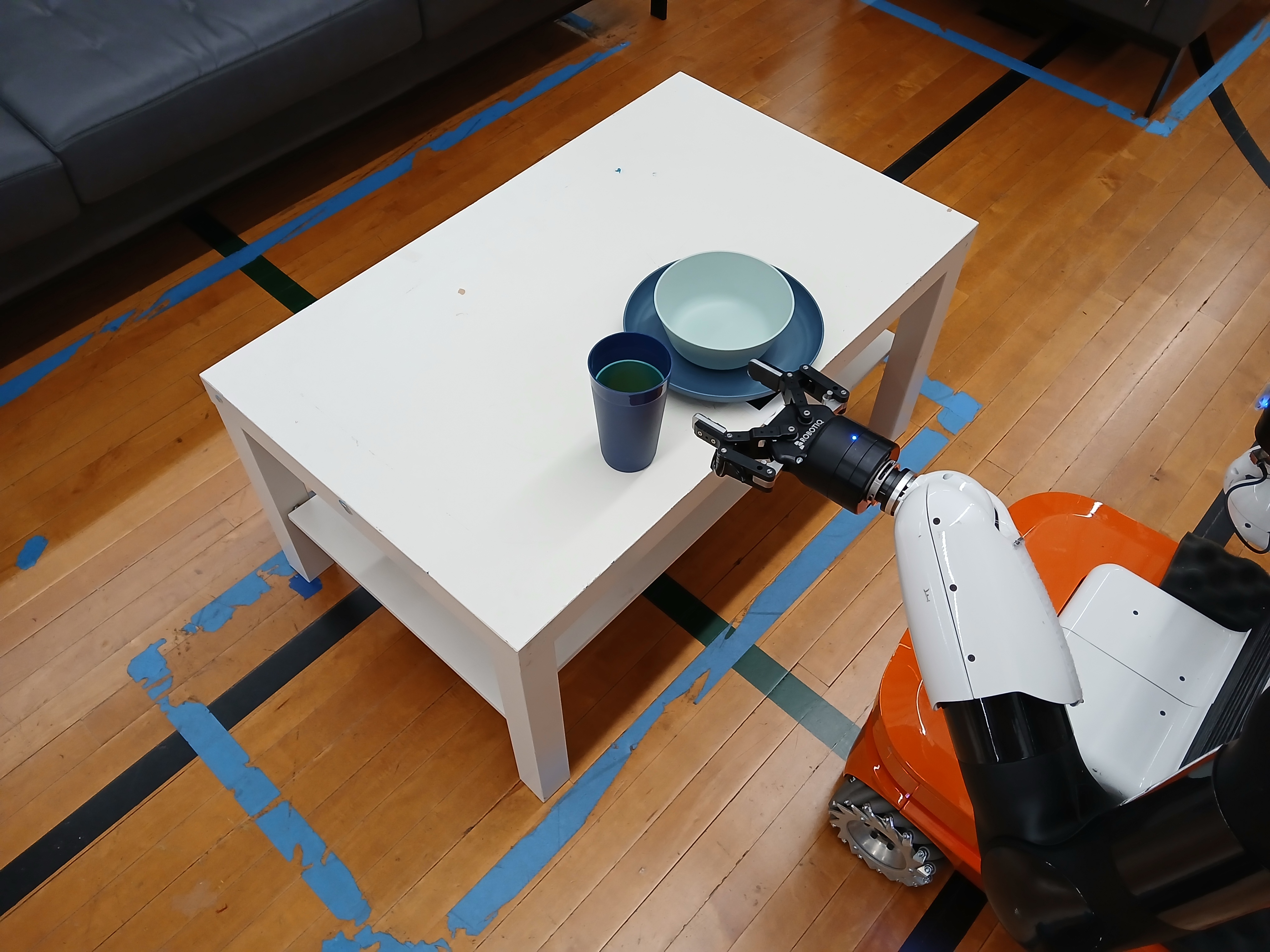}%
\hfill%
\includegraphics[width=0.32\linewidth]{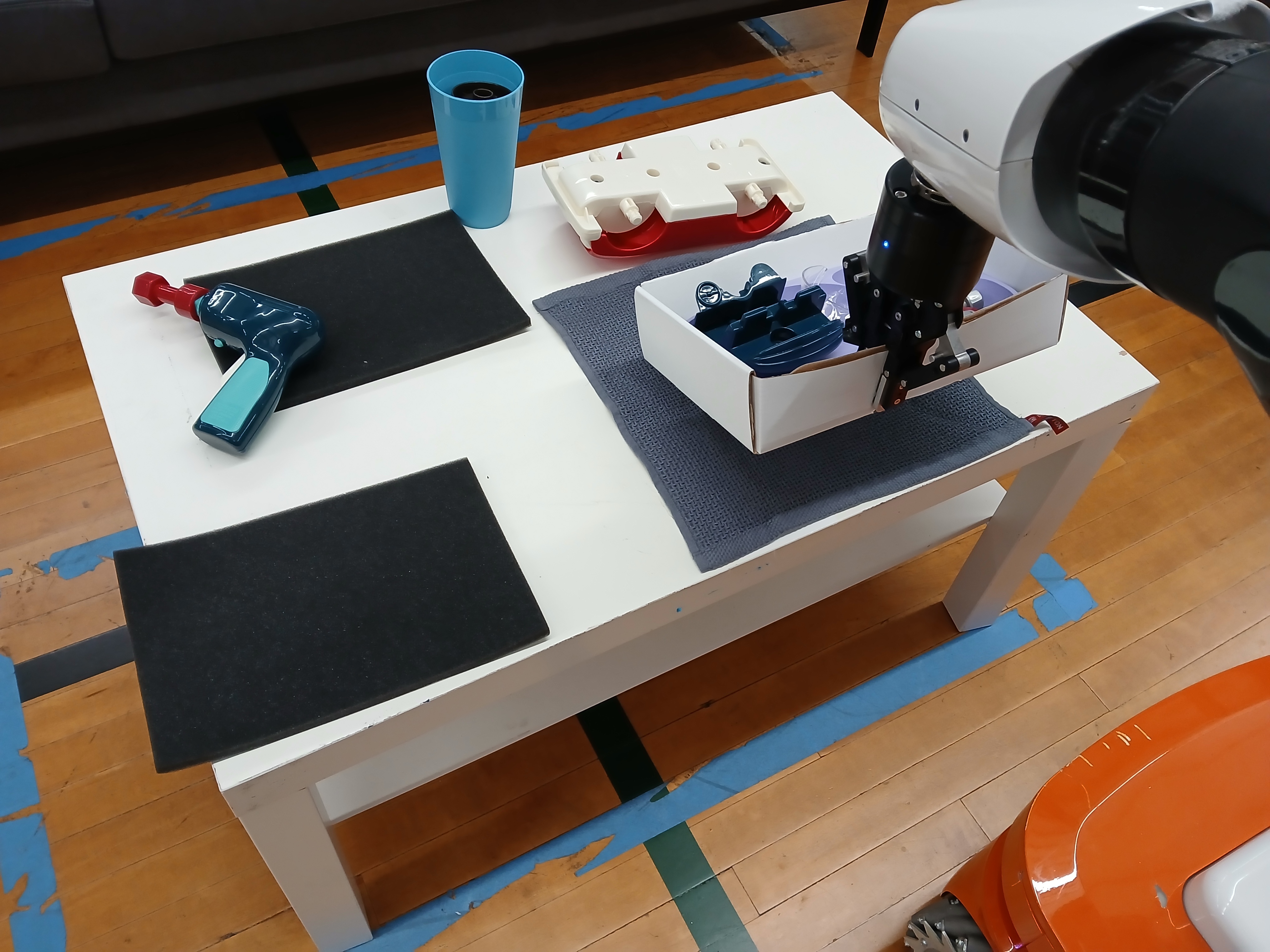}%
\hfill%
\includegraphics[width=0.32\linewidth]{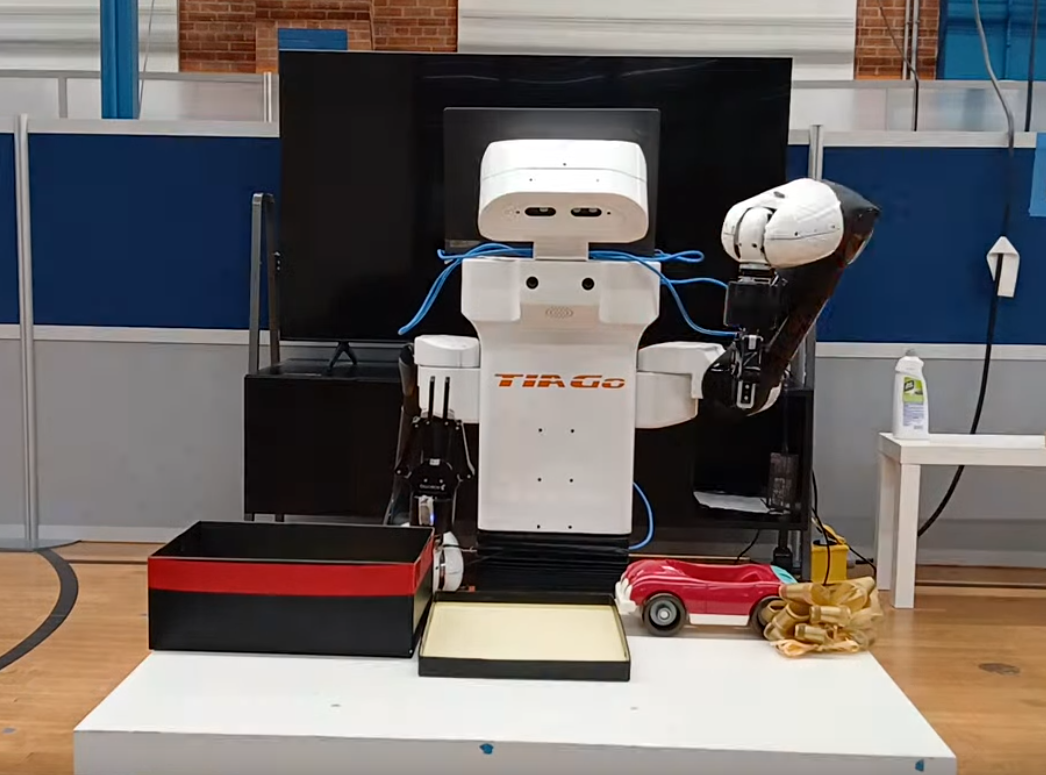}\\
\caption{Real-world tasks from left to right: pouring package into bowl, assembling toy car, and packing gift box.}  
\label{fig:rw-tasks}

\end{figure}

\subsection{Real-world Task Descriptions}
\label{app:task-description}

\subsection*{Task Plans}
Our real-world tasks are depicted in Fig.~\ref{fig:rw-tasks}. In \textbf{Task 1: Pour Package into Bowl}, the plan includes (steps 1-3) bringing the package, scissors, and bowl from the kitchen to the coffee table, (step 4) opening the package with the scissors, and (step 5) pouring the opened package into the bowl. The robot is incapable of performing step 4 and must rely on human help. 
In \textbf{Task 2: Assemble Toy Car}, the plan includes (steps 1-3) bringing the parts tray, drill, and wheels from the shelf to the coffee table, (step 4) using the drill and wheel caps from the parts tray to put the wheels onto the chassis, (steps 5-6) finding and switching the drill bit, and (steps 7-8) screwing in the window and seats onto the car with the drill. The robot is incapable of performing steps 4, 6, 7, 8, and has a low success rate for step 5.
In \textbf{Task 3: Pack Gift Box}, the plan includes (step 1) folding down the gift box flap, (steps 2-3) putting the tissue paper and toy car into the box, (steps 4-6) putting on the lid, getting the ribbons from the console table, and wrapping them around the box, and (steps 7-8) cutting a piece of tape to stick the gift bow to the top of the gift box. The robot is incapable of steps 4, 6, and 7 and has a low success rate for steps 2 and 5.

The minimum human effort required to complete the tasks ranged from just one step in Task 1 to four steps in Task 2, enabling us to test how our system compares with baselines in various regimes of dependence on human collaboration.

\subsection*{Hierarchical Plan Trees for Each Task}
The robot assumes the human only knows the high-level plan. It communicates about low-level steps only when necessary, such as to split up a high-level step. These are the high and low-level step breakdowns for each task, which we call the plan hierarchy. The low-level steps are listed here in skill-parameter pair format.

Task 1: Pour Package into Bowl (5 low-level steps)
\begin{enumerate}
    \item Bring bowl and package to coffee table.
        \begin{enumerate}
            \item \texttt{pickplace(bowl, coffee\_table)}
            \item \texttt{pickplace(package, coffee\_table)}
        \end{enumerate}
    \item Open package.
        \begin{enumerate}
            \item \texttt{pickplace(scissors, coffee\_table)}
            \item \texttt{pick\_open\_place(scissors, package, coffee\_table)}
        \end{enumerate}
    \item Pour package into bowl.
        \begin{enumerate}
            \item \texttt{pick\_pour\_place(package, bowl, coffee\_table)}
        \end{enumerate}
\end{enumerate}
Task 2: Assemble Toy Car (8 low-level steps)
\begin{enumerate}
    \item Bring parts to coffee table.
        \begin{enumerate}
            \item \texttt{pickplace(parts\_tray, coffee\_table)}
            \item \texttt{pickplace(wheels, coffee\_table)}
        \end{enumerate}
    \item Assemble wheels.
        \begin{enumerate}
            \item \texttt{pickplace(drill, coffee\_table)}
            \item \texttt{put\_on(wheels, car, drill)}
        \end{enumerate}
    \item Switch drill bit.
        \begin{enumerate}
            \item \texttt{pickplace(hex\_drill\_bit, coffee\_table)}
            \item \texttt{switch(hex\_drill\_bit, drill)}
        \end{enumerate}
    \item Assemble rest of car.
        \begin{enumerate}
            \item \texttt{put\_on(window, car, drill)}
            \item \texttt{put\_on(seats, car, drill)}
        \end{enumerate}
\end{enumerate}
Task 3: Pack Gift Box (8 low-level steps)
\begin{enumerate}
    \item Assemble box.
        \begin{enumerate}
            \item \texttt{fold(box\_flap)}
        \end{enumerate}
    \item Put in gift.
        \begin{enumerate}
            \item \texttt{pickplace(gift\_tissue\_paper, box)}
            \item \texttt{pickplace(toy\_car, box)}
        \end{enumerate}
    \item Seal the box.
        \begin{enumerate}
            \item \texttt{cover(box\_lid, box)}
            \item \texttt{pickplace(ribbons, coffee\_table)}
            \item \texttt{wrap(ribbons, box)}
        \end{enumerate}
    \item Decorate the box.
        \begin{enumerate}
            \item \texttt{cut\_put(tape, scissors, box)}
            \item \texttt{pickplace(gift\_bow, box\_lid)}
        \end{enumerate}
\end{enumerate}

\subsection{Dialog Excerpts from our User Studies}
\label{app:dialog-excerpt}

\subsubsection{Convincing a Reluctant User to Perform a Step}
Our system flexibly handles long dialog interactions like the following in Task 1 (opening and pouring the package into the bowl), where it manages to convince the human to help with opening the package.

\begin{dialogue}
\speak{Robot}{Could you please open the package using the scissors on the coffee table? Thank you so much!}
\speak{Human}{There are no scissors on the coffee table.}
\speak{Robot}{Let's collaborate to open the package! I can place the scissors onto the coffee table, and you can then use them to open the package. Thank you for your help!}
\speak{Human}{I think you should bring the scissors, and then you should use the scissors to open the package.}
\speak{Robot}{I'm happy to place the scissors on the coffee table for you. However, I'm not able to open the package with the scissors and place everything on the table. Let me know if there's anything I can do to assist.}
\speak{Human}{Ok, if you bring the scissors, then I will open the package with the scissors.}
\speak{Robot}{Great! I'll happily place the scissors on the coffee table. Thank you for taking care of opening the package with the scissors!}
\speaknoquote{Robot}{[brings the scissors]}
\speak{Human}{Thanks for bringing the scissors!}
\speak{Robot}{Could you please open the package with scissors and place it on the coffee table? Thank you!}
\speak{Human}{Ok, I will do that now!}
\speaknoquote{Human}{[opens the package]}
\end{dialogue}

\subsubsection{Failure Case}
In this example, \ours{} is unable to act on human-provided motion-level feedback (making the tissue paper ``lay flatter''). Additionally, \ours{} misunderstands the human and thinks they are talking about the ribbons step when the human is still talking about the tissue paper.
\begin{dialogue}
\speak{Human}{Please pack the gift.}
\speaknoquote{Robot}{[Puts tissue paper into the box.]}
\speak{Human}{It would be nice if the parchment was laying flatter.}
\speak{Human}{The compartment is kind of squished underneath the car. If you can hold the car for me, I can fix the parchment quickly.}
\speak{Robot}{I'm sorry, but I can't cover the box with the lid. Could you help with that part? Thank you so much! Could you kindly help with sealing the box? Specifically, could you cover the box with the lid? Thank you so much for your assistance!}
\speak{Human}{Yeah, I can, but can we fix the parchment first? If you can pick up the toy car, I can fix the parchment.}
\speak{Robot}{I'm delighted to hear you'll help! Unfortunately, I won't be able to wrap the ribbons around the box.}
\speak{Human}{Okay, that's fine. We can get to that later. Let me put the lid to the box on and then we can talk about the next steps.}
\end{dialogue}

See our website for the \href{https://robin-lab.cs.utexas.edu/MicoBot/user_study_dialog.html}{complete set of dialogs and operations in our human trials}. Contrast them with the \href{https://robin-lab.cs.utexas.edu/MicoBot/baseline_dialog.html}{dialog of our user studies with the LLM baseline}, which exhibit considerably less mixed-initiative dialog and collaborative success.

\subsection{\ours{} Implementation Details}
\label{app:ours-details}
\subsection*{Robot Q-function $Q_{R}$ training in OmniGibson}

\begin{figure}[h!]
  \centering

  \begin{subfigure}[b]{0.49\textwidth}
    \centering
    \includegraphics[width=0.495\textwidth]{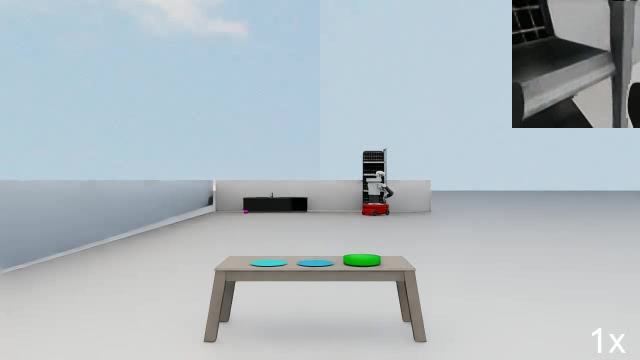}%
    \includegraphics[width=0.495\textwidth]{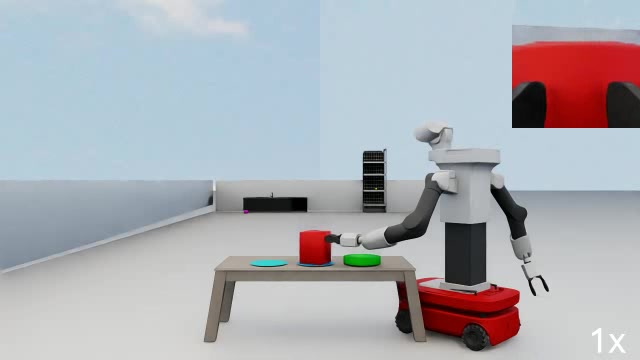}
    \caption*{Task 1: Pick package from shelf (left) and place on coffee table (right).}
  \end{subfigure}
  \hfill
  \begin{subfigure}[b]{0.49\textwidth}
    \centering
    \includegraphics[width=0.495\textwidth]{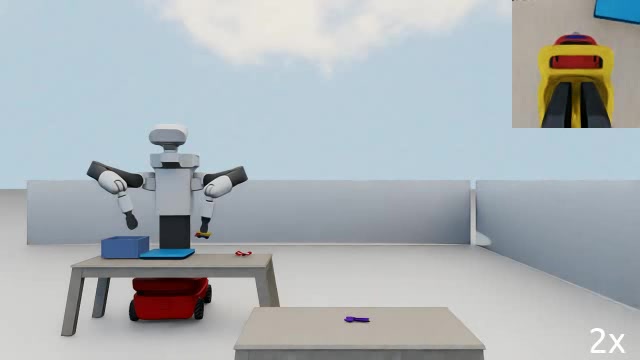}%
    \includegraphics[width=0.495\textwidth]{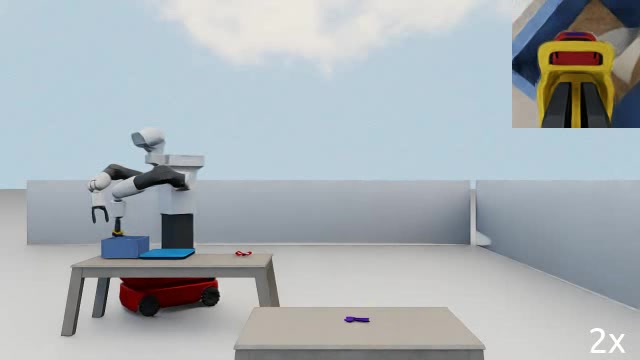}
    \caption*{Task 3: Pick toy car from coffee table (left) and place into gift box (right).}
  \end{subfigure}

  \caption{Frames from primitive rollouts in OmniGibson for task 1 (left two images) and task 3 (right two images). Left and right images within each task are frames near the beginning and end, respectively, of each skill. The square image at the top right of each frame represents the robot's camera view observation.}
  \label{fig:omnigibson}
\end{figure}

To train Q-functions for the robot, we first create a simulated OmniGibson environment with a PAL Tiago robot and an environment that roughly matches the relative locations of the relevant furnitures and objects. We then implemented each real-world skill first in OmniGibson.
Fig.~\ref{fig:omnigibson} depicts example frames from primitives in task 1 and task 3 we ran in the OmniGibson simulator to collect sample Q-values for each skill.

We collected samples of the form $(o, a, \mathcal{T})$, where $o$ is the initial observation of the world, $a$ is the skill-parameter pair $(\omega, \theta)$ taken by the robot at $o$, and $\mathcal{T}$ is the number of timesteps the robot takes to succeed at $a$ from $o$. If the robot does not succeed in its execution, then $\mathcal{T}$ is set to some fixed constant representing the maximum number of timesteps allowed in each skill-parameter execution.

To train our Q-functions, we collect roughly 100 samples for each action $a$ and train with inputs $(o, a)$ and target Q-values $-\mathcal{T}$ using $\ell_2$ regression with the Adam Optimizer. Since our observations $o$ are primarily symbolic but include some positional information of the robot and objects, our network architecture is extremely lightweight---2 linear layers with hidden size 32 and an output size of dimension 1 for the Q-value.

\subsection*{Human Q-function $Q_{H}$ Estimation}
To estimate $Q_{H}$, we compute two terms. The first is the human's stationary cost---the number of seconds for the human to perform some task if the relevant items were all right in front of them, within grasp. This term was copied from the output of an LLM call, which was prompted with a natural language description of the low-level step in the task alongside a URL to a product description of the toy car (for task 2). The second term is the human's traveling time---the number of seconds for the human to move from their current location to the object location. This was a simple 2D euclidean distance (in meters) between the assumed human location on the couch (in the real-world user studies) and the location of the objects, divided by the average human walking speed of $1.4 m/s$. We recognize this is a crude estimate of human effort, and we discuss the limitations of this in Section~\ref{sec:limitations-future-work}.

\subsection*{Forward Dynamics Model}
Our Q-functions rely on state and action inputs. However, computing the best task allocation involves considering Q-values for future steps, which depends on having knowledge of what the future state at that step will be. This involves creating a forward dynamics model so that we can estimate the future state $n$ plan steps into the future, which can be difficult to learn accurately for continuous states. We sidestep this problem by using symbolic states for our Q-values trained in simulation and maintaining these symbolic states during our real-world experiments. A symbolic state-based forward model is feasible to hardcode in our problem setting because we assume that each action affecting change in the world is a skill-parameter physical primitive, where the effect is quite easy to specify symbolically. For instance, the effect of \texttt{pickplace(bowl, coffee\_table)} is that the bowl moves from its original furniture to the coffee table. Though this is a limitation of our method, learning a forward dynamics model is not a contribution of our work, so we leave the extension of our approach to continuous state representations to future work.

\subsection{Additional Real-world Baselines and Success Rate / Human Effort Efficiency Metrics}
\label{app:more-rw-baselines-h-effort-efficiency}

In Table~\ref{tab:more-rw-baselines-effort-efficiency}, we compute two more oracle baselines based on our existing real-world experiments that help us understand the performance of \ours{} and the LLM-baseline: (B1) LLM baseline + \textit{oracle} skills + \textit{oracle} human (100\% helpful), and (B2) \textit{Oracle} task allocator (Robot performs all steps it has $>0\%$ success rate on) + real-world skills + \textit{oracle} human. While the real-world LLM baseline achieved 27.8\% success rate (as seen in Table~\ref{tab:real-world}), (B1) achieved 33.3\%, suggesting that the \textbf{LLM baseline was slightly hindered by primitive failures.} Even with an oracle task allocation, (B2) achieves only $44\%$ success, underperforming our method at $77.8\%$, demonstrating the importance of our method optimizing for task completion while minimizing human effort.

We also compute average human effort (seconds) and success rate per second of human effort. Our method uses human effort nearly as efficiently (0.83) as the oracle baselines (B1, 0.78; B2, 1.0). While the LLM baseline performs the best along this metric at 1.19, its advantage is primarily achieved solely through successes on the first task. On the two more temporally-extended tasks, our method matches or outperforms all baselines in success rate per unit of human effort.

\begin{table}[ht!] 
\caption{Additional Real-world Baselines and Human Effort Efficiency}
\label{tab:more-rw-baselines-effort-efficiency}
\centering
\resizebox{\columnwidth}{!}{ 
\begin{tabular}{rlcccc}
\toprule
\textbf{Real-World Task} &
\textbf{Metric} &
\textbf{LLM baseline} &
\textbf{\ours{}} &
\textbf{(B1)} &
\textbf{(B2)} \\
\midrule
\multirow{4}{*}{\textbf{Pour Package in Bowl}} & Success Rate (\%, $\uparrow$) & 83 & \textbf{100} & 83 & 41 \\
& Steps Completed (\%, $\uparrow$) & 93.3 & \textbf{100} & 93.3 & 64.5 \\
& Human Effort (seconds) & 36.7 & 48 & 46.7 & 32.7 \\
& Success Rate (\%) / Human Effort (s) ($\uparrow$) & \textbf{2.3} & 2.1 & 1.8 & 1.3 \\
\midrule
\multirow{4}{*}{\textbf{Assemble Toy Car}} & Success Rate (\%, $\uparrow$) & 0 & \textbf{67} & 0 & 11 \\
& Steps Completed (\%, $\uparrow$) & 29 & \textbf{94} & 54.2 & 43.2 \\
& Human Effort (seconds) & 20.8 & 197.3 & 56.7 & 55.0 \\
& Success Rate (\%) / Human Effort (s) ($\uparrow$) & 0 & \textbf{0.3} & 0 & 0.2 \\
\midrule
\multirow{4}{*}{\textbf{Pack Gift Box}} & Success Rate (\%, $\uparrow$) & 0 & 67 & 17 & \textbf{80} \\
& Steps Completed (\%, $\uparrow$) & 50 & 88 & 62.5 & \textbf{97.5} \\
& Human Effort (seconds) & 12.5 & 37.5 & 25 & 45.0 \\
& Success Rate (\%) / Human Effort (s) ($\uparrow$) & 0 & \textbf{1.8} & 0.68 & \textbf{1.8} \\
\midrule
\multirow{4}{*}{\textbf{Average}} & Success Rate (\%, $\uparrow$) & 27.8 & \textbf{77.8} & 33.3 & 44.0 \\
& Steps Completed (\%, $\uparrow$) & 58.2 & \textbf{93.8} & 70 & 68.4 \\
& Human Effort (seconds) & 23.3 & 94.3 & 42.8 & 44.2 \\
& Success Rate (\%) / Human Effort (s) ($\uparrow$) & \textbf{1.19} & 0.83 & 0.78 & 1.00 \\
\bottomrule
\end{tabular}
} 
\vspace{-1em}
\label{tab:new-results}
\end{table}

\subsection{Statistical Testing}
In Table~\ref{tab:stat-tests}, we perform statistical tests on our user study results from Table~\ref{tab:real-world}. All results are statistically significant ($p$-val. column).

\begin{table}[ht!]
\caption{Statistical Testing on Results Shown in Table~\ref{tab:real-world}}
\label{tab:stat-tests}
\centering
\resizebox{\columnwidth}{!}{
\begin{tabular}{lccccc}
\toprule
\textbf{Metric} &
\textbf{\ours{} (ours)} &
\textbf{LLM baseline} &
\textbf{Statistical Test} &
\textbf{Test statistic} &
\textbf{$p$-val.} \\
\midrule
Overall User Satisfaction ($\uparrow$, /5) & $3.61 \pm 0.95$ & $2.5 \pm 1.38$ & \multirow{4}{*}{\begin{minipage}{2cm}
    Wilcoxon \\ Signed-Rank
  \end{minipage}} & $W=17.0$ & $0.023540$ \\
Communicative Ability ($\uparrow$, /5) & $3.72 \pm 1.41$ & $2.33 \pm 1.56$ & & $W=8.5$ & $0.009233$ \\
Asked for Suitable Amt. of Help ($\uparrow$, /5) & $3.61 \pm 0.83$ & $2.56 \pm 1.42$ & & $W=15.0$ & $0.008636$ \\
Awareness of Its Limitations ($\uparrow$, /5) & $3.94 \pm 0.85$ & $2.61 \pm 1.53$ & & $W=16.0$ & $0.006502$ \\
\midrule
Success Rate (\%, $\uparrow$) & $77.8 \pm 15.7$ & $27.8 \pm 39.3$ & Fisher's Exact & -- & $0.006709$ \\
\midrule
Steps Completed (\%, $\uparrow$) & $93.8 \pm 15.2$ & $57.5 \pm 30.6$ & Wilcoxon Signed-Rank & $W=0.0$ & $0.002031$ \\
\bottomrule
\end{tabular}
} 
\label{tab:performance_comparison}
\end{table}

\subsection{Mixed-Initiative Dialog: Real-world Metrics}
\label{app:mixed-init-dialog-metrics}

\begin{table}[h]
\caption{Mixed-Initiative Dialog Metrics}
\label{tab:mixed-init-dialog-metrics}
\centering
\resizebox{\columnwidth}{!}{
\begin{tabular}{lcc}
\toprule
\begin{tabular}{@{}c@{}}\textbf{Metric (avg over each trial)} \\ R = Robot, H = human \end{tabular} &
\textbf{\ours{} (ours)} &
\textbf{LLM baseline} \\
\midrule
\# R-Helpreqs & $2.9 \pm 1.4$ & $0.9 \pm 1.1$ \\
Initial H acceptance rate & $55\% \pm 35\%$ & $70\% \pm 46\%$ \\
H acceptance rate after R-negotiation & $86\% \pm 27\%$ & $75\% \pm 40\%$ \\
R-init dialogs & $3.6 \pm 1.8$ & $0.9\pm 1.1$ \\
H-init dialogs & $2.9 \pm 2.4$ & $2.3 \pm 2.9$ \\
Initiative Shifts & $2.4 \pm 2.1$ & $1.1 \pm 1.6$ \\
\bottomrule
\end{tabular}
} 
\label{tab:dialog_stats}
\end{table}

We evaluate the dialog of all our real-world user studies across the three tasks and compile mixed-initiative metrics of \ours{} and the baseline in Table~\ref{tab:dialog_stats}.
\textbf{\ours{} boosts human acceptance of help requests from $\mathbf{55\%}$ to $\mathbf{86\%}$ with negotiation.} However, the LLM baseline, which succeeded at a far lower rate in our user studies, made far fewer requests (0.9 vs. \ours{}'s 2.9) and achieves a smaller acceptance increase ($70\%$ to $75\%$). \textbf{\ours{} collaborates with a high level of robot and human initiated dialog ($\mathbf{3.6}$ robot dialog initiations vs. $\mathbf{2.9}$ human initiations, with $\mathbf{2.4}$ dialog initiative shifts/trial}), whereas the LLM trials are human-initiative driven ($0.9$ robot dialog initiations vs. $2.3$ humaan initiations, with $0.9$ initiative shifts). This suggests that mixed-initiative dialog helps enable \ours{} to have better task success outcomes and user satisfaction ratings than the LLM baseline.

\subsection{Detailed Simulation Results}
\label{app:ours-sim-results}
\subsubsection{Setup}
In simulation, we ran our method, the three baselines (RL, LLM, random), and our method's four ablations (no $p_{H, t}$ estimation, no plan hierarchy, no R-initiative dialog, and no H-initiative dialog) on eight different settings of parameterized humans in simulation. These eight settings were a cross product of 2 dialog mood settings (positive and negative) and 4 ground-truth $\tilde{p}_{H, t} \in \{0.0, 0.3, 0.7, 1.0\}$ settings (following the notation introduced in Appendix~\ref{app:extra-exps}, where the $\tilde{p}$ denotes the ground truth probability while the plain $p$ denotes our estimate). $10$ trials were run for each method in each of the eight settings for the parameterized human.

\subsubsection{Simulation Experiments}

In Table~\ref{tab:sim-results}, we show the results of our method in a simulation version of our real-world Task 1. Our method performs better than baselines especially in scenarios where $\tilde{p}_{H, t}$ is low, because our method is able to take initiative through dialog, such as by proposing ways to split up steps to make them more achievable with the simulated human. The averages in Table~\ref{tab:sim-results} are plotted in Fig.~\ref{fig:rw-success-vs-human-effort}.

\begin{table*}[h]
\centering
\caption{Simulation Task 1 Performance across different $\tilde{p}_{H, t}$ Values and Language Sentiments.}
\label{tab:sim-results}
\resizebox{\textwidth}{!}{
\begin{tabular}{|l|l|c|c|c|c|c|c|c|c|c|}
\cline{3-10}
\multicolumn{2}{c|}{} & \multicolumn{8}{c|}{\textbf{Human Parameters (Mood, $\tilde{p}_{H, t}$)}} & \multicolumn{1}{c}{} \\
\hline
\multirow{2}{*}{\textbf{Method}} & \multirow{2}{*}{\textbf{Metric}} 
& \multicolumn{4}{c|}{\textbf{Positive Mood}} 
& \multicolumn{4}{c|}{\textbf{Negative Mood}} 
& \multirow{2}{*}{\textbf{Avg. (\%)}} \\
\cline{3-10}
 &  & \textbf{0.0} & \textbf{0.3} & \textbf{0.7} & \textbf{1.0} 
 & \textbf{0.0} & \textbf{0.3} & \textbf{0.7} & \textbf{1.0} & \\
\hline
\multirow{3}{*}{\textbf{Ours}} 
& Success Rate & 3/10 & 6/10 & 9/10 & 10/10 & 1/10 & 4/10 & 9/10 & 9/10 & \textbf{63.75} \\
& Num Plan Steps Completed & 3.6/5 & 4.2/5 & 4.8/5 & 5.0/5 & 3.2/5 & 3.8/5 & 4.8/5 & 4.5/5 & \textbf{84.5} \\
& Prop. Plan Steps done by Human & 0.1667 & 0.2381 & 0.3125 & 0.4 & 0.03125 & 0.1579 & 0.354 & 0.377 & 25.47 \\
\hline
\multirow{3}{*}{\textbf{LLM Baseline}} 
& Success Rate & 2/10 & 2/10 & 4/10 & 7/10 & 3/10 & 6/10 & 6/10 & 6/10 & 45 \\
& Num Plan Steps Completed & 3.4/5 & 3.4/5 & 3.7/5 & 4.4/5 & 3.6/5 & 4.2/5 & 4.0/5 & 4.2/5 & 77.25 \\
& Prop. Plan Steps done by Human & 0.0588 & 0.05882 & 0.2162 & 0.1591 & 0.1111 & 0.1428 & 0.175 & 0.166 & 13.6 \\
\hline
\multirow{3}{*}{\textbf{Random Agent}} 
& Success Rate & 2/10 & 5/10 & 6/10 & 7/10 & 2/10 & 3/10 & 6/10 & 7/10 & 47.5 \\
& Num Plan Steps Completed & 3.4/5 & 3.5/5 & 4.0/5 & 4.4/5 & 3.4/5 & 2.8/5 & 4.0/5 & 4.4/5 & 74.75 \\
& Prop. Plan Steps done by Human & 0.1176 & 0.4286 & 0.525 & 0.7045 & 0.1176 & 0.2143 & 0.525 & 0.7045 & 41.71 \\
\hline
\multirow{3}{*}{\textbf{RL}} 
& Success Rate &  0/10 & 1/10 & 4/10 & 10/10 & 0/10 & 1/10 & 4/10 & 10/10 & 37.5 \\
& Num Plan Steps Completed & 2.4/5 & 2.3/5 & 3.4/5 & 5.0/5 & 2.4/5 & 2.3/5 & 3.4/5 & 5.0/5 & 65.5 \\
& Prop. Plan Steps done by Human & 0.125 & 0.1739 & 0.4412 & 0.54 & 0.125 & 0.1739 & 0.4412 & 0.54 & 32.0 \\
\hline
\multirow{3}{*}{\textbf{Only R Init}} 
& Success Rate & 0/10 & 3/10 & 9/10 & 10/10 & 0/10 & 5/10 & 9/10 & 10/10 & 57.5 \\
& Num Plan Steps Completed & 3.0/5 & 3.6/5 & 4.8/5 & 5.0/5 & 3.0/5 & 4.0/5 & 4.8/5 & 5.0/5 & 83 \\
& Prop. Plan Steps done by Human & 0.0 & 0.1111 & 0.3542 & 0.4 & 0.0 & 0.225 & 0.354 & 0.4 & 23.05 \\
\hline
\multirow{3}{*}{\textbf{Only H Init}} 
& Success Rate & 0/10 & 0/10 & 0/10 & 0/10 & 2/10 & 0/10 & 0/10 & 2/10 & 5.0 \\
& Num Plan Steps Completed & 3.0/5 & 3.0/5 & 3.0/5 & 3.0/5 & 3.2/5 & 3.0/5 & 3.0/5 & 3.3/5 & 61.25 \\
& Prop. Plan Steps done by Human & 0.0 & 0.0 & 0.0 & 0.0/3.0 & 0.1875 & 0.0 & 0.0 & 0.1212 & 3.86 \\
\hline
\multirow{3}{*}{\textbf{Ours w/o p\_help}} 
& Success Rate & 3/10 & 5/10 & 9/10 & 10/10 & 2/10 & 3/10 & 9/10 & 9/10 & 62.5 \\
& Num Plan Steps Completed & 3.6/5 & 4.0/5 & 4.8/5 & 5.0/5 & 3.4/5 & 3.4/5 & 4.8/5 & 4.7/5 & 84.25 \\
& Prop. Plan Steps done by Human & 0.1667 & 0.3 & 0.3333 & 0.38 & 0.1176 & 0.2059 & 0.3125 & 0.4468 & 28.29 \\
\hline
\multirow{3}{*}{\textbf{Ours w/o Plan Hier.}} 
& Success Rate & 2/10 & 4/10 & 7/10 & 10/10 & 0/10 & 3/10 & 4/10 & 8/10 & 47.5 \\
& Num Plan Steps Completed & 3.4/5 & 3.8/5 & 4.0/5 & 5.0/5 & 3.0/5 & 3.4/5 & 3.6/5 & 4.2/5 & 76 \\
& Prop. Plan Steps done by Human & 0.0588 & 0.1316 & 0.25 & 0.24 & 0.0667 & 0.1176 & 0.1944 & 0.2381 & 16.22 \\
\hline
\end{tabular}
}
\end{table*}

\subsection{User Study Details}
\label{app:user-study-details}
\subsubsection{User Instructions}
Users were read the following instructions at the beginning of the study. (Instructions here are shown for Task 2.)

\begin{enumerate}
    \item Thank you so much for coming for our user study! We wanted to remind you to review the RIS before proceeding, and that you may voluntarily opt-out of the study at any time.
    \item You are working with the robot to perform the task of assembling the toy car. You must use the hexagonal drill bit to screw in the wheels, and the phillips drill bit to screw in the seat and the window. [Demonstrate these steps to the human]. You and the robot operate on a shared understanding of the plan. [Read the 4 high-level steps of the plan tree for this task. Do NOT discuss the low-level steps of the plan tree.]
    \item Our goal is to simulate a home robot setting, where the human (you) is relatively busy with your own tasks, and once in a while you provide physical assistance and talk to the robot. So you are free to do work during each trial.
    \item Once the robot asks you to do a step, and you accept, you must finish that step successfully.
    \item We will perform 2 trials, each of a different method.
    \item Both you and the robot can do a subset of the steps in the plan. You will communicate with the robot to determine who does what steps.
    \item These are the objects you will work with during the task. I will move them now to their initial positions where they will start at the beginning of each trial. [Move objects to initial positions.]
    \item For safety, I will gate-keep each of the robot’s physical actions. In other words, the actions are generated by the robot itself, but they will be displayed on the laptop screen with a confirmation message, and I can either allow that physical action to be executed by the robot, or block the action from being executed if it brings the robot to an unsafe location.
    \item The robot will stay on the TV side of the coffee table, while you will sit on the couch and stay on the couch side of the coffee table.
    \item You are free to get up off the couch if you want to volunteer to perform steps that involve going to the sink or shelf, but you can only go when the robot is stationary and waiting on the other side of the coffee table. Steps are done in sequential order; our system doesn’t support parallelization (agents working simultaneously).
    \item You will be communicating to the robot through this headset. We will perform a mic-check now to make sure it can pick up your voice. [Do mic check.]
    \item Now, this is what the robot will sound like when it talkes to you. [Play audio sample of the robot.] Try responding to it, and I will see if it can hear you.
    \item The systems today can handle different kinds of dialog. (1) refusal/acceptance, (2) task allocation, such as (``Could you pour the package in the plate later?'' Or: ``I can pour the package onto the plate later.''), (3) silence---you don't need to respond to the robot every time, and (4) a proposal to split up adjacent steps, such as ``Please bring me the drill so that I can put on the wheels.'' You may engage in any of these types of dialog, and the robot may also engage in them when communicating to you.
    \item Do you have any questions before we start? I will let you know when each trial begins and ends. Sometimes trials may end prematurely.
\end{enumerate}

\subsubsection{Success Rate}
Success at each step is measured by whether the goal state of a primitive has been achieved. For instance, a \texttt{pickplace(obj, furniture)} step in the plan is marked as successfully completed if the \texttt{obj} ends up on the \texttt{furniture} after execution. This means that primitive errors (such as a pickplace operation that accidentally moves the object off of the furniture as the arm is retracting) count as a failed execution. In Table~\ref{tab:real-world}, ``\% of task steps completed'' is evaluated by tallying up all of the steps in the low-level plan that have succeeded and dividing this by the length of the plan. ``Entire Task Success Rate'' is defined as whether all plan steps in a run have been successfully completed.

\subsection{Real-world Failure Analysis}
\begin{figure}[ht!]
    \centering
    \includegraphics[width=1\columnwidth]{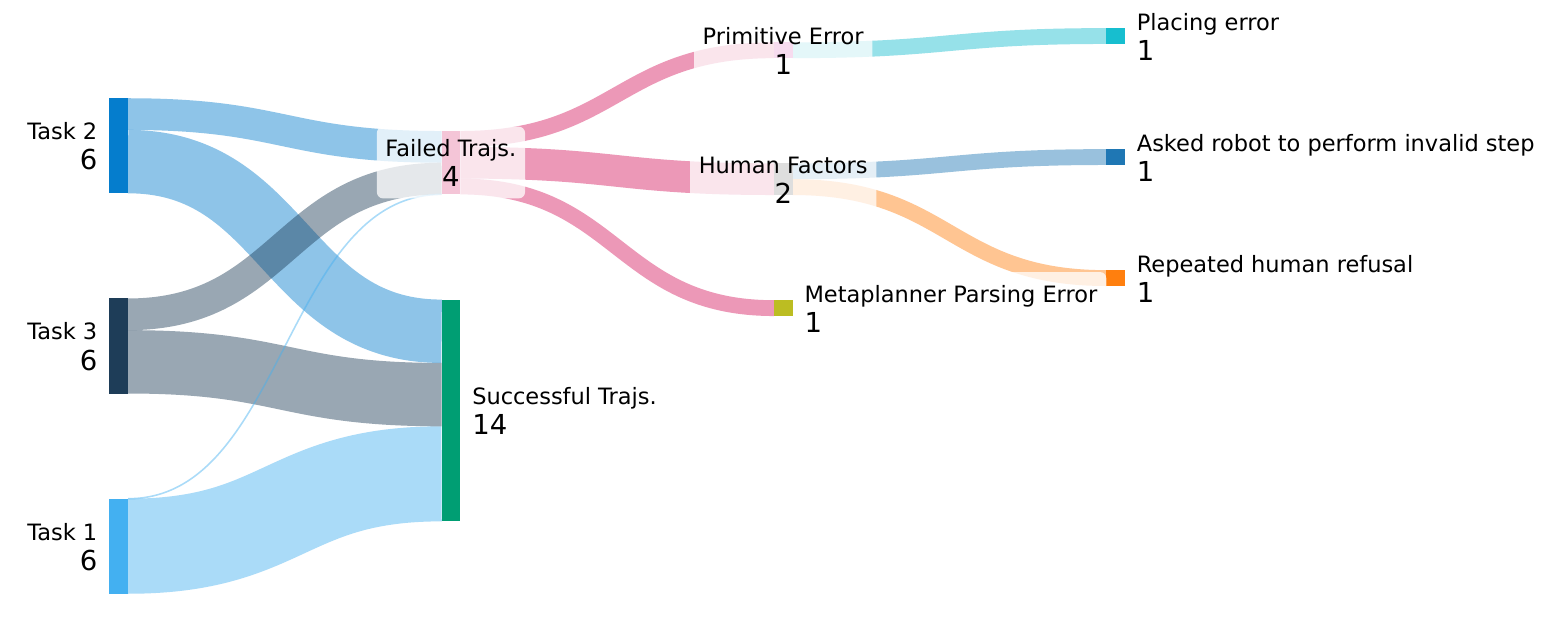}
    \caption{\ours{} mainly fails from an inability to solicit human help. Metaplanner and primitive errors were also encountered.}
    \label{fig:ours-failure-flow-diagram}
\end{figure}

\begin{figure}[ht!]
    \centering
    \includegraphics[width=1\columnwidth]{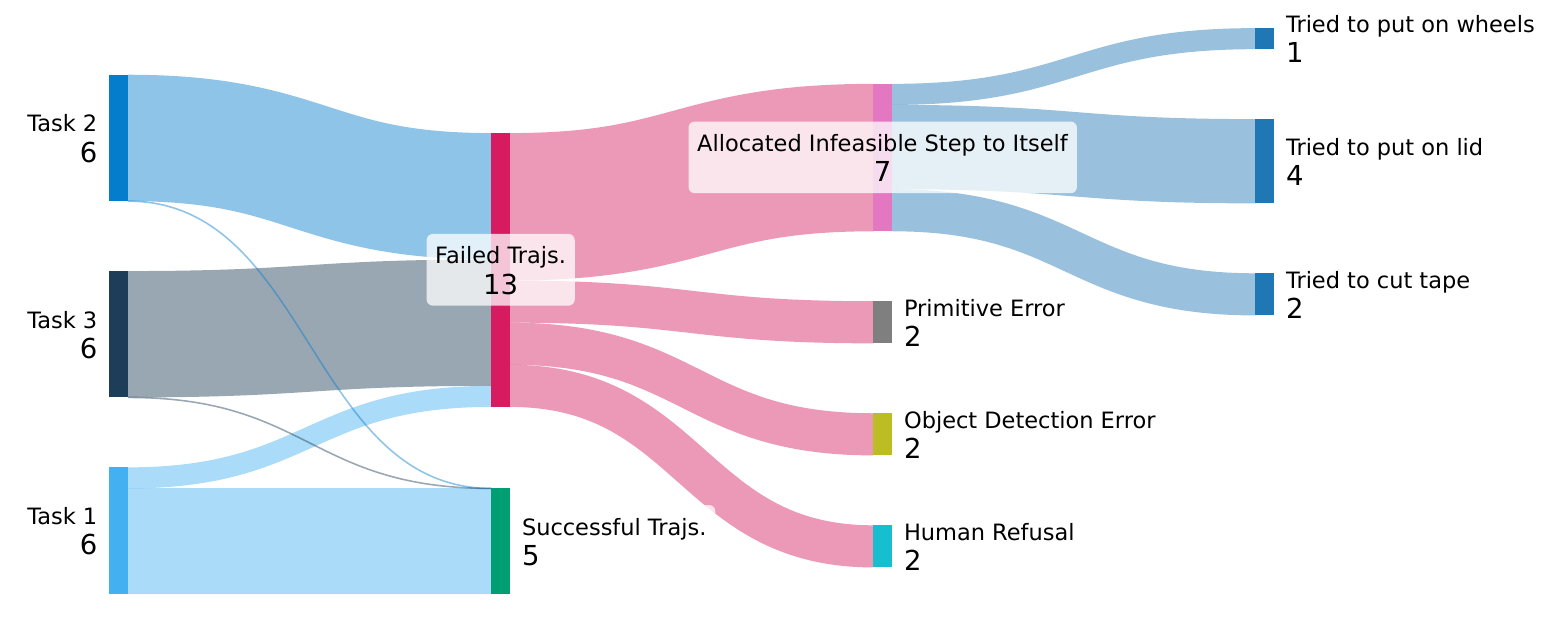}
    \caption{The LLM baseline failed mainly by allocating itself steps it could not perform, due to the LLM's lack of knowledge of its own affordances.}
    \label{fig:llm-baseline-failure-flow-diagram}
\end{figure}

See the failure breakdowns in our real-world trials for \ours{} (Figure~\ref{fig:ours-failure-flow-diagram}) and the LLM baseline (Figure~\ref{fig:llm-baseline-failure-flow-diagram}).

\begin{enumerate}
    \item Task 1: Cut and Pour Package into Bowl. 0 failed trials out of 6.
    \item Task 2: Assemble Toy Car. 2 failed trials out of 6. 1 failure that was rectified by human.
        \begin{itemize}
            \item 1 dialog error: user wanted robot to perform a non-step plan outside of its capabilities, and refused when robot said it wasn't able to perform it
            \item 1 primitive error: robot did not release its grasp of the drill.
            \item 1 metaplanner dialog parsing error
        \end{itemize}
    \item Task 3: Pack Gift Box. 2 failed trials out of 6.
        \begin{itemize}
            \item 1 primitive error: placed bow on box but bow dropped to floor as gripper retracted
            \item 1 termination condition triggered: human rejected robot's help request and subsequent proposals twice in a row.
        \end{itemize}
\end{enumerate}

\begin{enumerate}
    \item Task 1: Cut and Pour Package into Bowl. 1 failed trials out of 6.
        \begin{itemize}
            \item 1 termination condition triggered: human rejected robot's help request twice in a row.
        \end{itemize}
    \item Task 2: Assemble Toy Car. 6 failed trials out of 6.
        \begin{itemize}
            \item 2 primitive errors: Placed the drill, but then it fell off of the table.
            \item 2 perception errors: Unable to visually find the correct location to place the object.
            \item 1 task allocation error: Allocated to put on the car wheels itself.
            \item 1 termination condition triggered: Fell into a conversational loop with the user.
        \end{itemize}
    \item Task 3: Pack Gift Box. 6 failed trials out of 6. 1 failure was rectified by human.
        \begin{itemize}
            \item 6 task allocation errors: Tried to put on the lid itself four times; tried to cut a piece of tape itself twice.
            \item 1 primitive error: Inadvertently dropped the car onto the floor as it was trying to place it.
        \end{itemize}
\end{enumerate}

\subsection{Fault Recovery}
\label{app:fault-recovery}
The metaplanner ocassionally produces faulty, non-executable code. For fault recovery, the metaplanner is automatically re-queried up to 2 additional times to create code. If these attempts also produce non-executable code, the most recent dialog from the human is ignored for 2 further, automated metaplanner requeries. These re-queries are handled by a try-except block in the iterative planner module of \ours{}.

\subsection{RL Baseline Details}
\label{app:rl-baseline}
For our RL baseline which was evaluated in simulation, we train a hierarchical policy where the high-level policy was a task allocator that outputted logits over two classes: 0 (Robot would perform current step), or 1 (Human would perform current step). If the logit for 0 is higher, then the image observation is passed into the low-level robot policy that decides the discrete physical action to take in the world. Otherwise, the robot asks the human through a verbal action for help on that step. We use sparse rewards, issued only when all 5 steps were completed in the task, in the proper order.

We initially trained the RL policy on two simulated human settings: one where the human ground truth $\tilde{p}_{H, t} = 1.0$, and another where $\tilde{p}_{H, t} \sim U[0, 1]$. We were unable to obtain policies with any non-zero training returns after thousands of iterations on the latter setting, so we only report results on the former setting, which explains why the RL policy does not perform well when $\tilde{p}_{H, t}$ is low.

\subsection{Additional Experimental Investigations}
\label{app:extra-exps}
In addition to those discussed in Section~\ref{sec:exp}, we explore the following additional experimental questions.

\textbf{(4) How important is $p_{H, t}$ estimation at adapting to human collaborators?}
A correct estimation of the true likeliness of a human to help, $\tilde{p}_{H, t}$, is critical: overestimation causes \methodname{} to overly rely on human effort, potentially decreasing user satisfaction, while underestimation lowers task success outcomes if the robot needs to rely on its low-success-rate skills instead of asking the human for help. 

\begin{figure*}[t!]
    \centering
    \includegraphics[width=1\textwidth,trim={0 0.0cm 0 0},clip]{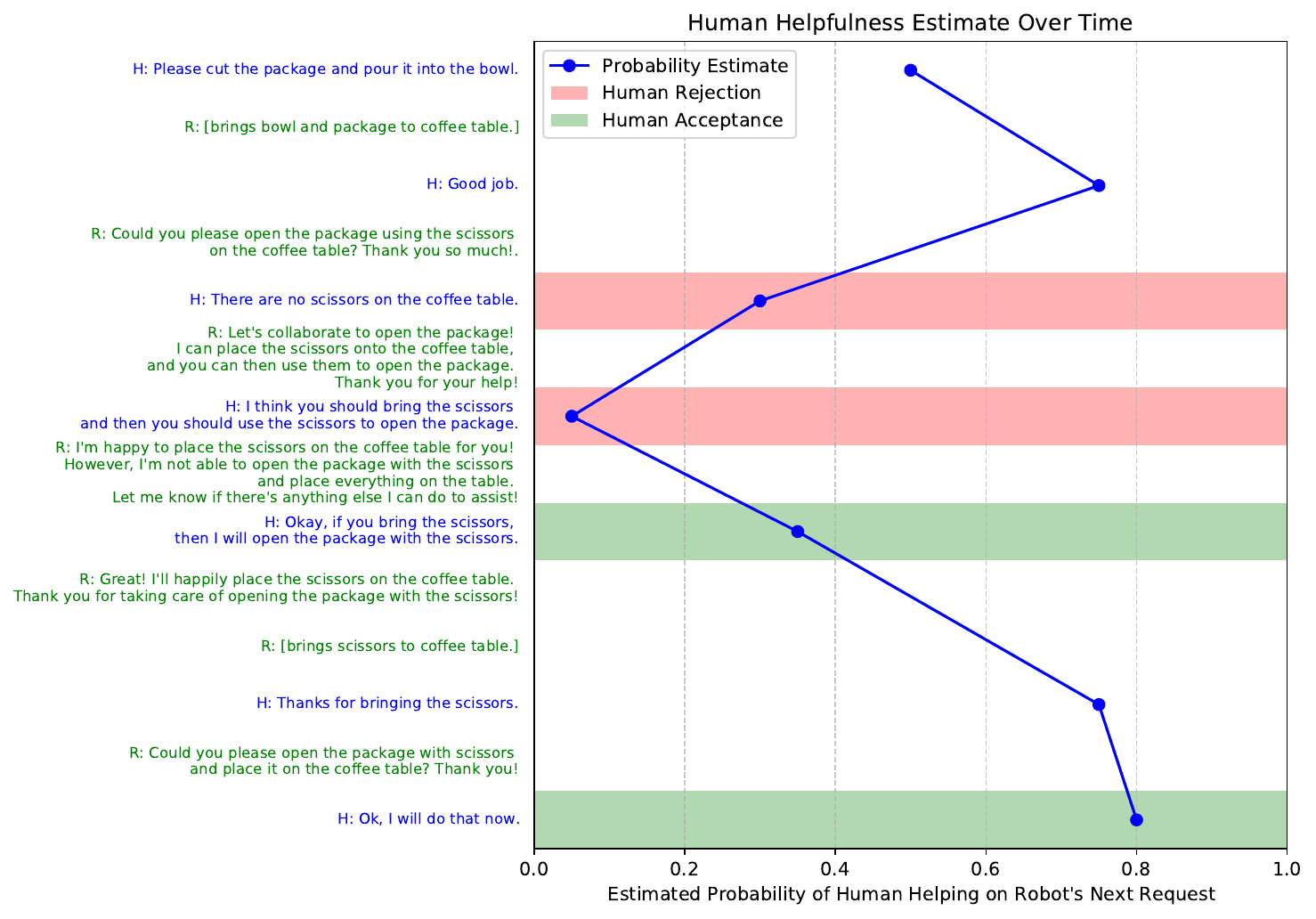}
    \caption{From a real-world user study: \ours{}'s $p_{H, t}$ estimation (blue line) reacts in real time to the human's rejections (red), acceptances (green), and encouraging remarks. All dialog is shown as $y$-labels. Green text denotes robot actions/dialog, and blue text denotes human dialog. The timestep $t$ increases from top to bottom on the $y$-axis. 
    }
    \label{fig:p-help-plot-trial}
\end{figure*}

First, we examine in Fig.~\ref{fig:p-help-plot-trial} a real-world instance of how well \ours{} can estimate the probability of the human helping on the next turn during the course of a user study. After the robot's help request was rejected twice in a row (top 2 red horizontal stripes), the robot's helpfulness estimate of the human plummets to $0.05$ (5\% estimated likelihood of the human helping the robot). However, after the robot explains its incapacity to use scissors, the human accepts the next two help requests (in green) and the robot's helpfulness estimate of the human increases to $0.8$. Note that simple comments from the human, such as a ``Thank you'' or ``good job,'' also had positive effects on the estimated $p_{H, t}$, because the robot inferred that the human was in a more positive mood and hence more likely to help. This graph demonstrates that \ours{} is fairly competent at estimating a reasonable $p_{H, t}$ value when calculating the human q-values for each step in the plan.

To analyze the effect of a good $p_{H, t}$ estimate on task allocation, we demonstrate through a controlled toy-setting in simulation in Table~\ref{tab:sim-alloc-switch} exactly how the optimal task allocation changes as the robot discovers more information about the human's willingness to help. Steps that are optimally allocated to the human are shown in \colorbox{LightBlue}{blue}, and steps optimally allocated to the robot are shown in \colorbox{LightGreen}{green}. The Q-values of the selected agent in each cell are shown in parentheses. Table~\ref{tab:sim-alloc-switch} depicts a rollout on the open and pour package into bowl (Task 1) in simulation, which has the same 5 step plan as the real-world Task 1 described in \ref{app:task-description}. Unlike our real-world experiments, where $\alpha=10$, in Table~\ref{tab:sim-alloc-switch}) we set $\alpha=0.3$ for illustrative purposes, which sets human effort to be around $3\times$ \textit{cheaper} than robot effort. In this toy setting, we program the human to reject the robot's first help request but to help the robot when it asks a second time.

Initially ($t=0$) all steps are allocated to the human. When the human rejects the initial help request from the robot, the $p_{H, t}$ estimate drops to $0.25$, increasing the Q-values of the human and switching the allocation of all but steps 2-3 to the robot after just two environment timesteps ($t=2$). (Recall that the robot cannot perform step 3, and due to the hierarchical structure of our plan, steps 2 and 3 are bundled together as an abstract step.) This demonstrates that having a good $p_{H, t}$ estimate is crucial to adapt to the human's willingness to help. Since the human demonstrated initial unwillingness to help, \ours{} quickly learns to decrease its $p_{H, t}$ estimate and allocate many more steps to itself by the second timestep. Had \ours{} not properly estimated $p_{H, t}$, the robot would have repeatedly asked the human for help even if the human was extremely unwilling to, leading to worse user satisfaction in working with the robot.

\begin{table}
\caption{Computed Best Task Allocation (and Agent Q-values) During a Sim Trial on Task 1.}
\label{tab:sim-alloc-switch}
\centering
\resizebox{\columnwidth}{!}{
\begin{tabular}{c|c|c|c|c|c}
Env. Timestep & Step 1 & Step 2 & Step 3 & Step 4 & Step 5 \\
\hline
$t=0$ & \cellcolor{LightBlue}H (-9.6) & \cellcolor{LightBlue}H (-7.2) & \cellcolor{LightBlue}H (-13.2) & \cellcolor{LightBlue}H (-2.4) & \cellcolor{LightBlue}H (-2.4) \\
$t=2$ & \cellcolor{LightGreen}R (-13.0) & \cellcolor{LightGreen}R (-9.0) & \cellcolor{LightBlue}H (-13.2) & \cellcolor{LightBlue}H (-4.8) & \cellcolor{LightGreen}R (-1.0) \\
$t=6$ & -- & \cellcolor{LightGreen}R (-12.0) & \cellcolor{LightBlue}H (-13.2) & \cellcolor{LightBlue}H (-4.8) & \cellcolor{LightGreen}R (-1.0) \\
$t=9$ & -- & -- & \cellcolor{LightBlue}H (-13.2) & \cellcolor{LightBlue}H (-4.8) & \cellcolor{LightGreen}R (-1.0) \\
$t=16$ & -- & -- & -- & -- & \cellcolor{LightGreen}R (-3.0) \\
\hline
\end{tabular}
}
\end{table}

\subsection{Meta-planner and $p_{H}$ Estimator Accuracy}
We tested LLM-generated meta-planner programs against manually-annotated ground truth in 6 user studies (3 successful + 3 failed rollouts; 59 programs total). The meta-planner achieved an \textbf{$\mathbf{89.8\%}$ accuracy (53/59 programs)}.

We also tested \ours{}'s accuracy at estimating the likeliness of human helping, $p_{H, t}$. Across 33 estimates from the same 6 user studies, \ours{}'s mean absolute error (MAE) against a ground-truth (proportion of human-accepted help requests so far in the trial) was 0.11 (on a scale of 0 to 1.0). \textbf{76\% of estimates were within 0.15 of the ground truth.}

\subsection{Further Connections to Prior Work}
\label{app:extra-related-work}
\subsubsection{Agents with Both Physical and Verbal Actions}
\methodname{} relies on a heterogeneous action space that includes interacting with the physical world \textit{and} generating freeform dialogue to a human collaborator. 
Prior works have developed \textbf{policies with a combined physical and verbal action space} through RL~\cite{Das2017ICCV, shervedani2024multimodalreinforcementlearningrobots}
or IL (imitation learning)~\cite{padmakumar2022teach, deepmindinteractiveagentsteam2022creatingmultimodalinteractiveagents}. 
Research on language emergence in multiagent systems~\cite{Peters_2025, kottur-etal-2017-natural} has also examined how cooperative agents learn to communicate through latent representations or natural language when performing simulated robotic tasks~\cite{lin2021learninggroundmultiagentcommunication, lowe2020interactionsupervisionselfplayemergent, woodward2019learninginteractivelylearnassist, lobostsunekawa2022madreamercoordinationcommunicationshared, kolb2019learningrequestguidanceemergent}. However, these works are typically limited to simulated domains, where action spaces and task dynamics are highly abstracted or simplified. They often rely on limited communication protocols without integrating grounded task structure, rich human preferences, or real-world execution constraints.
In contrast, \methodname{} leverages an LLM to generate freeform, grounded dialog within a shared task context, enabling fluid mixed-initiative interaction and reasoning over both verbal and physical actions in real-world scenarios.

\subsubsection{Natural Language and Robotics}
Our work sits at the broad, growing intersection of natural language and robot learning. We refer the reader to various lines of work upon which different modules of our method are based, including language-conditioned robot policies~\cite{jang2021bc, lynch2021language, calvin2021, hulc22, shao2020concept, sodhani2021multi, silva2021langcon, karamcheti2021lila, jiang2022vima, shah2023mutex, yu2022deltaco}, LLMs as task planners~\cite{huang2022inner, saycan2022arxiv, chen2022nlmapsaycan, raman2023cape}, code-based policies~\cite{codeaspolicies2022, pmlrv235li24ar, huang2023voxposercomposable3dvalue}, hierarchical policies~\cite{shi2025hirobotopenendedinstruction, shi2024yellrobotimprovingonthefly, belkhale2024rthactionhierarchiesusing} and planners~\cite{choi2025reactree, luo2023obtaining}, vision-language representations~\cite{radford2021learning, zhai2021lit, zhu2023languagebind} for robotic control~\cite{nair2022r3m, shridhar2021cliport, shridhar2022peract, yu2024lang4sim2real}, and language-based reward shaping for RL policies~\cite{nair2021lorel, goyal2019learn, goyal2020pixl2r, fan2022minedojo, ma2023liv, ma2024eurekahumanlevelrewarddesign, ma2024dreureka, yu2023languagerewardsroboticskill}.



\section*{Acknowledgments}
We thank all user study participants and the lab members of the RobIn Lab and Machine Learning Lab at UT Austin for their feedback, suggestions, and discussion during the course of this work. We also thank Rutav Shah and Arpit Bahety for their help in debugging real-world robot issues, and Ben Abbatematteo for his edits on an early copy of the manuscript. Finally, we thank the anonymous reviewers for their constructive feedback. This research was funded in part by DARPA TIAMAT HR0011-24-9-0428, Amazon Award, Emerson Electric, and the UT Austin Graduate School Dissertation Writing Fellowship.

\printbibliography

@misc{calvin2021,
  doi = {10.48550/ARXIV.2112.03227},
  
  url = {https://arxiv.org/abs/2112.03227},
  
  author = {Mees, Oier and Hermann, Lukas and Rosete-Beas, Erick and Burgard, Wolfram},
  
  title = {CALVIN: A Benchmark for Language-conditioned Policy Learning for Long-horizon Robot Manipulation Tasks},
  
  publisher = {arXiv},
  
  year = {2021},
  
  copyright = {arXiv.org perpetual, non-exclusive license}
}

@inproceedings{goyal2020pixl2r,
  url = {https://arxiv.org/abs/2007.15543},
  author = {Prasoon Goyal and Scott Niekum and Raymond Mooney},
  title = {PixL2R: Guiding Reinforcement Learning Using Natural Language by Mapping Pixels to Rewards},
  journal = {4th Annual Conference on Robot Learning},
  year = {2020},
}

@inproceedings{goyal2019learn,
  url = {https://arxiv.org/abs/1903.02020},
  author = {Prasoon Goyal and Scott Niekum and Raymond Mooney},
  title = {Using Natural Language for Reward Shaping in Reinforcement Learning},
  journal = {IJCAI},
  year = {2019},
}

@inproceedings{jang2021bc,
    title={{BC}-Z: Zero-Shot Task Generalization with Robotic Imitation Learning},
    author={Eric Jang and Alex Irpan and Mohi Khansari and Daniel Kappler and Frederik Ebert and Corey Lynch and Sergey Levine and Chelsea Finn},
    booktitle={5th Annual Conference on Robot Learning},
    year={2021},
    url={https://openreview.net/forum?id=8kbp23tSGYv}
}

@article{lynch2021language,
  title   = {Language Conditioned Imitation Learning over Unstructured Data},
  author  = {Lynch, Corey and Sermanet, Pierre},
  journal = {Robotics: Science and Systems},
  year    = {2021},
  url     = {https://arxiv.org/abs/2005.07648},
}

@inproceedings{
    nair2021lorel,
    title={Learning Language-conditioned Robot Behavior from Offline Data and Crowd-Sourced Annotation},
    author={Suraj Nair and Eric Mitchell and Kevin Chen and Brian Ichter and Silvio Savarese and Chelsea Finn},
    booktitle={5th Annual Conference on Robot Learning },
    year={2021},
    url={https://arxiv.org/pdf/2109.01115}
}

@inproceedings{shridhar2021cliport,
  title     = {CLIPort: What and Where Pathways for Robotic Manipulation},
  author    = {Shridhar, Mohit and Manuelli, Lucas and Fox, Dieter},
  booktitle = {Proceedings of the 5th Conference on Robot Learning (CoRL)},
  year      = {2021},
}

@article{radford2021learning,
  title={Learning transferable visual models from natural language supervision},
  author={Radford, Alec and Kim, Jong Wook and Hallacy, Chris and Ramesh, Aditya and Goh, Gabriel and Agarwal, Sandhini and Sastry, Girish and Askell, Amanda and Mishkin, Pamela and Clark, Jack and others},
  journal={arXiv preprint arXiv:2103.00020},
  year={2021}
}

@inproceedings{
    karamcheti2021lila,
    title={LILA: Language-Informed Latent Actions},
    author={Siddharth Karamcheti and Megha Srivastava and Percy Liang and Dorsa Sadigh},
    booktitle={5th Annual Conference on Robot Learning },
    year={2021},
    url={https://arxiv.org/pdf/2111.03205}
}

@article{hulc22,
    author = {Oier Mees and Lukas Hermann and Wolfram Burgard},
    title = {What Matters in Language Conditioned Imitation Learning},
    journal={arXiv preprint arXiv:2204.06252},
    year = 2022,
}

@article{nair2022r3m,
    author = {Suraj Nair and Aravind Rajeswaran and Vikash Kumar and Chelsea Finn and Abhinav Gupta},
    title = {R3M: A Universal Visual Representation for Robot Manipulation},
    journal={arXiv preprint arXiv:2203.12601},
    year = 2022,
}

@article{sodhani2021multi,
  title={Multi-Task Reinforcement Learning with Context-based Representations},
  author={Sodhani, Shagun and Zhang, Amy and Pineau, Joelle},
  journal={arXiv preprint arXiv:2102.06177},
  year={2021}
}

@article{shridhar2022peract,
  title={Perceiver-Actor: A Multi-Task Transformer for Robotic Manipulation},
  author={Mohit Shridhar and Lucas Manuelli and Dieter Fox},
  journal={Conference on Robot Learning},
  year={2022}
}

@inproceedings{huang2022inner,
    title={Inner Monologue: Embodied Reasoning through Planning with Language Models},
    author={Wenlong Huang and Fei Xia and Ted Xiao and Harris Chan and Jacky Liang and Pete Florence and Andy Zeng and Jonathan Tompson and Igor Mordatch and Yevgen Chebotar and Pierre Sermanet and Noah Brown and Tomas Jackson and Linda Luu and Sergey Levine and Karol Hausman and Brian Ichter},
    booktitle={arXiv preprint arXiv:2207.05608},
    year={2022}
}

@inproceedings{saycan2022arxiv,
    title={Do As {I} Can and Not As {I} Say: Grounding Language in Robotic Affordances},
    author={Michael Ahn and Anthony Brohan and Noah Brown and Yevgen Chebotar and Omar Cortes and Byron David and Chelsea Finn and Chuyuan Fu and Keerthana Gopalakrishnan and Karol Hausman and Alex Herzog and Daniel Ho and Jasmine Hsu and Julian Ibarz and Brian Ichter and Alex Irpan and Eric Jang and Rosario Jauregui Ruano and Kyle Jeffrey and Sally Jesmonth and Nikhil Joshi and Ryan Julian and Dmitry Kalashnikov and Yuheng Kuang and Kuang-Huei Lee and Sergey Levine and Yao Lu and Linda Luu and Carolina Parada and Peter Pastor and Jornell Quiambao and Kanishka Rao and Jarek Rettinghouse and Diego Reyes and Pierre Sermanet and Nicolas Sievers and Clayton Tan and Alexander Toshev and Vincent Vanhoucke and Fei Xia and Ted Xiao and Peng Xu and Sichun Xu and Mengyuan Yan and Andy Zeng},
    booktitle={arXiv preprint arXiv:2204.01691},
    year={2022}
}

@inproceedings{shao2020concept,
 title={Concept2Robot: Learning Manipulation Concepts from Instructions and Human Demonstrations},
 author={Shao, Lin and Migimatsu, Toki and Zhang, Qiang and Yang, Karen and Bohg, Jeannette},
 booktitle={Proceedings of Robotics: Science and Systems (RSS)},
 year={2020},
}

@inproceedings{chen2022nlmapsaycan,
    title={Open-vocabulary Queryable Scene Representations for Real World Planning},
    author={Boyuan Chen and Fei Xia and Brian Ichter and Kanishka Rao and Keerthana Gopalakrishnan and Michael S. Ryoo and Austin Stone and Daniel Kappler},
    booktitle={arXiv preprint arXiv:2209.09874},
    year={2022}
}

@inproceedings{silva2021langcon,
  title={LANCON-LEARN: Learning with Language to Enable Generalization in Multi-Task Manipulation},
  author={Andrew Silva and Nina Moorman and William Silva and Zulfiqar Zaidi and Nakul Gopalan and Matthew Gombolay},
  booktitle={IEEE Robotics and Automation Letters},
  year={2021}
}

@article{zhu2023languagebind,
  title={Languagebind: Extending video-language pretraining to n-modality by language-based semantic alignment},
  author={Zhu, Bin and Lin, Bin and Ning, Munan and Yan, Yang and Cui, Jiaxi and Wang, HongFa and Pang, Yatian and Jiang, Wenhao and Zhang, Junwu and Li, Zongwei and others},
  journal={arXiv preprint arXiv:2310.01852},
  year={2023}
}

@article{fan2022minedojo,
  title={Minedojo: Building open-ended embodied agents with internet-scale knowledge},
  author={Fan, Linxi and Wang, Guanzhi and Jiang, Yunfan and Mandlekar, Ajay and Yang, Yuncong and Zhu, Haoyi and Tang, Andrew and Huang, De-An and Zhu, Yuke and Anandkumar, Anima},
  journal={Advances in Neural Information Processing Systems},
  volume={35},
  pages={18343--18362},
  year={2022}
}

@article{jiang2022vima,
  title={Vima: General robot manipulation with multimodal prompts},
  author={Jiang, Yunfan and Gupta, Agrim and Zhang, Zichen and Wang, Guanzhi and Dou, Yongqiang and Chen, Yanjun and Fei-Fei, Li and Anandkumar, Anima and Zhu, Yuke and Fan, Linxi},
  journal={arXiv},
  year={2022}
}

@article{shah2023mutex,
  title={MUTEX: Learning unified policies from multimodal task specifications},
  author={Shah, Rutav and Mart{\'\i}n-Mart{\'\i}n, Roberto and Zhu, Yuke},
  journal={arXiv preprint arXiv:2309.14320},
  year={2023}
}

@article{yu2022deltaco,
  title={Using both demonstrations and language instructions to efficiently learn robotic tasks},
  author={Yu, Albert and Mooney, Raymond J},
  journal={arXiv preprint arXiv:2210.04476},
  year={2022}
}

@inproceedings{raman2023cape,
  title={Cape: Corrective actions from precondition errors using large language models},
  author={Raman, Shreyas Sundara and Cohen, Vanya and Paulius, David and Idrees, Ifrah and Rosen, Eric and Mooney, Ray and Tellex, Stefanie},
  booktitle={2nd Workshop on Language and Robot Learning: Language as Grounding},
  year={2023}
}

@article{ma2023liv,
  title={LIV: Language-Image Representations and Rewards for Robotic Control},
  author={Ma, Yecheng Jason and Liang, William and Som, Vaidehi and Kumar, Vikash and Zhang, Amy and Bastani, Osbert and Jayaraman, Dinesh},
  journal={arXiv preprint arXiv:2306.00958},
  year={2023}
}

@misc{liu2023grounding,
      title={Grounding DINO: Marrying DINO with Grounded Pre-Training for Open-Set Object Detection}, 
      author={Shilong Liu and Zhaoyang Zeng and Tianhe Ren and Feng Li and Hao Zhang and Jie Yang and Chunyuan Li and Jianwei Yang and Hang Su and Jun Zhu and Lei Zhang},
      year={2023},
      eprint={2303.05499},
      archivePrefix={arXiv},
      primaryClass={cs.CV}
}

@article{zhai2021lit,
  author       = {Xiaohua Zhai and
                  Xiao Wang and
                  Basil Mustafa and
                  Andreas Steiner and
                  Daniel Keysers and
                  Alexander Kolesnikov and
                  Lucas Beyer},
  title        = {LiT: Zero-Shot Transfer with Locked-image Text Tuning},
  journal      = {CoRR},
  volume       = {abs/2111.07991},
  year         = {2021},
  url          = {https://arxiv.org/abs/2111.07991},
  eprinttype    = {arXiv},
  eprint       = {2111.07991},
  bibsource    = {dblp computer science bibliography, https://dblp.org}
}

@misc{yu2023promptbasedmontecarlotreesearch,
      title={Prompt-Based Monte-Carlo Tree Search for Goal-Oriented Dialogue Policy Planning}, 
      author={Xiao Yu and Maximillian Chen and Zhou Yu},
      year={2023},
      eprint={2305.13660},
      archivePrefix={arXiv},
      primaryClass={cs.CL},
      url={https://arxiv.org/abs/2305.13660}, 
}

@misc{qian2021databasesearchresultsdisambiguation,
      title={Database Search Results Disambiguation for Task-Oriented Dialog Systems}, 
      author={Kun Qian and Ahmad Beirami and Satwik Kottur and Shahin Shayandeh and Paul Crook and Alborz Geramifard and Zhou Yu and Chinnadhurai Sankar},
      year={2021},
      eprint={2112.08351},
      archivePrefix={arXiv},
      primaryClass={cs.CL},
      url={https://arxiv.org/abs/2112.08351}, 
}

@misc{deng2024plugandplaypolicyplannerlarge,
      title={Plug-and-Play Policy Planner for Large Language Model Powered Dialogue Agents}, 
      author={Yang Deng and Wenxuan Zhang and Wai Lam and See-Kiong Ng and Tat-Seng Chua},
      year={2024},
      eprint={2311.00262},
      archivePrefix={arXiv},
      primaryClass={cs.CL},
      url={https://arxiv.org/abs/2311.00262}, 
}

@misc{deng2023promptingevaluatinglargelanguage,
      title={Prompting and Evaluating Large Language Models for Proactive Dialogues: Clarification, Target-guided, and Non-collaboration}, 
      author={Yang Deng and Lizi Liao and Liang Chen and Hongru Wang and Wenqiang Lei and Tat-Seng Chua},
      year={2023},
      eprint={2305.13626},
      archivePrefix={arXiv},
      primaryClass={cs.CL},
      url={https://arxiv.org/abs/2305.13626}, 
}

@misc{chen2024learningclarifymultiturnconversations,
      title={Learning to Clarify: Multi-turn Conversations with Action-Based Contrastive Self-Training}, 
      author={Maximillian Chen and Ruoxi Sun and Sercan O. Arık and Tomas Pfister},
      year={2024},
      eprint={2406.00222},
      archivePrefix={arXiv},
      primaryClass={cs.CL},
      url={https://arxiv.org/abs/2406.00222}, 
}

@ARTICLE{796083,

  author={Allen, J.E. and Guinn, C.I. and Horvtz, E.},

  journal={IEEE Intelligent Systems and their Applications}, 

  title={Mixed-initiative interaction}, 

  year={1999},

  volume={14},

  number={5},

  pages={14-23},

  doi={10.1109/5254.796083}
}

@inproceedings{chu-carroll-2000-mimic,
    title = "{MIMIC}: An Adaptive Mixed Initiative Spoken Dialogue System for Information Queries",
    author = "Chu-Carroll, Jennifer",
    booktitle = "Sixth Applied NLP Conference",
    month = apr,
    year = "2000",
    address = "Seattle, Washington, USA",
    publisher = "ACL",
    url = "https://aclanthology.org/A00-1014/",
    doi = "10.3115/974147.974161",
    pages = "97--104"
}

@misc{wang2024mosaicmodularassistiveinteractive,
      title={MOSAIC: A Modular System for Assistive and Interactive Cooking}, 
      author={Huaxiaoyue Wang and Kushal Kedia and Juntao Ren and Rahma Abdullah and Atiksh Bhardwaj and Angela Chao and Kelly Y Chen and Nathaniel Chin and Prithwish Dan and Xinyi Fan and Gonzalo Gonzalez-Pumariega and Aditya Kompella and Maximus Adrian Pace and Yash Sharma and Xiangwan Sun and Neha Sunkara and Sanjiban Choudhury},
      year={2024},
      eprint={2402.18796},
      archivePrefix={CoRL},
      primaryClass={cs.RO},
      url={https://arxiv.org/abs/2402.18796}, 
}

@misc{mandi2023rocodialecticmultirobotcollaboration,
      title={RoCo: Dialectic Multi-Robot Collaboration with Large Language Models}, 
      author={Zhao Mandi and Shreeya Jain and Shuran Song},
      year={2023},
      eprint={2307.04738},
      archivePrefix={arXiv},
      primaryClass={cs.RO},
      url={https://arxiv.org/abs/2307.04738}, 
}

@misc{feng2024largelanguagemodelbasedhumanagent,
      title={Large Language Model-based Human-Agent Collaboration for Complex Task Solving}, 
      author={Xueyang Feng and Zhi-Yuan Chen and Yujia Qin and Yankai Lin and Xu Chen and Zhiyuan Liu and Ji-Rong Wen},
      year={2024},
      eprint={2402.12914},
      archivePrefix={arXiv},
      primaryClass={cs.CL},
      url={https://arxiv.org/abs/2402.12914}, 
}

@misc{ren2023robotsaskhelpuncertainty,
      title={Robots That Ask For Help: Uncertainty Alignment for Large Language Model Planners}, 
      author={Allen Z. Ren and Anushri Dixit and Alexandra Bodrova and Sumeet Singh and Stephen Tu and Noah Brown and Peng Xu and Leila Takayama and Fei Xia and Jake Varley and Zhenjia Xu and Dorsa Sadigh and Andy Zeng and Anirudha Majumdar},
      year={2023},
      eprint={2307.01928},
      archivePrefix={arXiv},
      primaryClass={cs.RO},
      url={https://arxiv.org/abs/2307.01928}, 
}

@misc{bennetot2020artificialagentsaskhelp,
      title={Should artificial agents ask for help in human-robot collaborative problem-solving?}, 
      author={Adrien Bennetot and Vicky Charisi and Natalia Díaz-Rodríguez},
      year={2020},
      eprint={2006.00882},
      archivePrefix={arXiv},
      primaryClass={cs.LG},
      url={https://arxiv.org/abs/2006.00882}, 
}

@misc{suhr2022executinginstructionssituatedcollaborative,
      title={Executing Instructions in Situated Collaborative Interactions}, 
      author={Alane Suhr and Claudia Yan and Charlotte Schluger and Stanley Yu and Hadi Khader and Marwa Mouallem and Iris Zhang and Yoav Artzi},
      year={2022},
      eprint={1910.03655},
      archivePrefix={arXiv},
      primaryClass={cs.CL},
      url={https://arxiv.org/abs/1910.03655}, 
}

@misc{kojima2021continuallearninggroundedinstruction,
      title={Continual Learning for Grounded Instruction Generation by Observing Human Following Behavior}, 
      author={Noriyuki Kojima and Alane Suhr and Yoav Artzi},
      year={2021},
      eprint={2108.04812},
      archivePrefix={arXiv},
      primaryClass={cs.CL},
      url={https://arxiv.org/abs/2108.04812}, 
}

@misc{deepmindinteractiveagentsteam2022creatingmultimodalinteractiveagents,
      title={Creating Multimodal Interactive Agents with Imitation and Self-Supervised Learning}, 
      author={DeepMind Interactive Agents Team and Josh Abramson and Arun Ahuja and Arthur Brussee and Federico Carnevale and Mary Cassin and Felix Fischer and Petko Georgiev and Alex Goldin and Mansi Gupta and Tim Harley and Felix Hill and Peter C Humphreys and Alden Hung and Jessica Landon and Timothy Lillicrap and Hamza Merzic and Alistair Muldal and Adam Santoro and Guy Scully and Tamara von Glehn and Greg Wayne and Nathaniel Wong and Chen Yan and Rui Zhu},
      year={2022},
      eprint={2112.03763},
      archivePrefix={arXiv},
      primaryClass={cs.LG},
      url={https://arxiv.org/abs/2112.03763}, 
}

@misc{gao2023alexaarenausercentricinteractive,
      title={Alexa Arena: A User-Centric Interactive Platform for Embodied AI}, 
      author={Qiaozi Gao and Govind Thattai and Suhaila Shakiah and Xiaofeng Gao and Shreyas Pansare and Vasu Sharma and Gaurav Sukhatme and Hangjie Shi and Bofei Yang and Desheng Zheng and Lucy Hu and Karthika Arumugam and Shui Hu and Matthew Wen and Dinakar Guthy and Cadence Chung and Rohan Khanna and Osman Ipek and Leslie Ball and Kate Bland and Heather Rocker and Yadunandana Rao and Michael Johnston and Reza Ghanadan and Arindam Mandal and Dilek Hakkani Tur and Prem Natarajan},
      year={2023},
      eprint={2303.01586},
      archivePrefix={arXiv},
      primaryClass={cs.HC},
      url={https://arxiv.org/abs/2303.01586}, 
}

@INPROCEEDINGS{7379306, author={Jiang, Shu and Arkin, Ronald C.}, booktitle={2015 IEEE International Conference on Systems, Man, and Cybernetics}, title={Mixed-Initiative Human-Robot Interaction: Definition, Taxonomy, and Survey}, year={2015}, volume={}, number={}, pages={954-961}, doi={10.1109/SMC.2015.174}}

@ARTICLE{4081977,

  author={Carbonell, Jaime R.},

  journal={IEEE Transactions on Man-Machine Systems}, 

  title={AI in CAI: An Artificial-Intelligence Approach to Computer-Assisted Instruction}, 

  year={1970},

  volume={11},

  number={4},

  pages={190-202},

  doi={10.1109/TMMS.1970.299942}}

@INPROCEEDINGS{4107804,

  author={Few, Douglas A. and Bruemmer, David J. and Walton, Miles C.},

  booktitle={ROMAN 2006 - The 15th IEEE International Symposium on Robot and Human Interactive Communication}, 

  title={Improved Human-Robot Teaming through Facilitated Initiative}, 

  year={2006},

  volume={},

  number={},

  pages={171-176},

  doi={10.1109/ROMAN.2006.314413}}

@misc{natarajan2024mixedinitiativehumanrobotteamingsuboptimality,
      title={Mixed-Initiative Human-Robot Teaming under Suboptimality with Online Bayesian Adaptation}, 
      author={Manisha Natarajan and Chunyue Xue and Sanne van Waveren and Karen Feigh and Matthew Gombolay},
      year={2024},
      eprint={2403.16178},
      archivePrefix={arXiv},
      primaryClass={cs.RO},
      url={https://arxiv.org/abs/2403.16178}, 
}

@misc{paleja2024designsenablingcollaborationhumanmachine,
      title={Designs for Enabling Collaboration in Human-Machine Teaming via Interactive and Explainable Systems}, 
      author={Rohan Paleja and Michael Munje and Kimberlee Chang and Reed Jensen and Matthew Gombolay},
      year={2024},
      eprint={2406.05003},
      archivePrefix={arXiv},
      primaryClass={cs.RO},
      url={https://arxiv.org/abs/2406.05003}, 
}

@INPROCEEDINGS{9106686,
  author={Bishop, Justin and Burgess, Jaylen and Ramos, Cooper and Driggs, Jade B. and Williams, Tom and Tossell, Chad C. and Phillips, Elizabeth and Shaw, Tyler H. and Visser, Ewart J. de},
  booktitle={2020 Systems and Information Engineering Design Symposium (SIEDS)}, 
  title={CHAOPT: A Testbed for Evaluating Human-Autonomy Team Collaboration Using the Video Game Overcooked!2}, 
  year={2020},
  volume={},
  number={},
  pages={1-6},
  doi={10.1109/SIEDS49339.2020.9106686}}

@misc{rosero2021cooksunderstandingdynamichumanagent,
      title={Two Many Cooks: Understanding Dynamic Human-Agent Team Communication and Perception Using Overcooked 2}, 
      author={Andres Rosero and Faustina Dinh and Ewart J. de Visser and Tyler Shaw and Elizabeth Phillips},
      year={2021},
      eprint={2110.03071},
      archivePrefix={arXiv},
      primaryClass={cs.HC},
      url={https://arxiv.org/abs/2110.03071}, 
}

@article{Faccio2024,
  author    = {Maurizio Faccio and Irene Granata and Riccardo Minto},
  title     = {Task allocation model for human-robot collaboration with variable cobot speed},
  journal   = {Journal of Intelligent Manufacturing},
  volume    = {35},
  pages     = {793--806},
  year      = {2024},
  doi       = {10.1007/s10845-023-02073-9},
  url       = {https://link.springer.com/article/10.1007/s10845-023-02073-9}
}

@misc{vats2022synergisticschedulinglearningallocation,
      title={Synergistic Scheduling of Learning and Allocation of Tasks in Human-Robot Teams}, 
      author={Shivam Vats and Oliver Kroemer and Maxim Likhachev},
      year={2022},
      eprint={2203.07478},
      archivePrefix={arXiv},
      primaryClass={cs.RO},
      url={https://arxiv.org/abs/2203.07478}, 
}

@article{YU2021487,
title = {Optimizing task scheduling in human-robot collaboration with deep multi-agent reinforcement learning},
journal = {Journal of Manufacturing Systems},
volume = {60},
pages = {487-499},
year = {2021},
issn = {0278-6125},
doi = {https://doi.org/10.1016/j.jmsy.2021.07.015},
url = {https://www.sciencedirect.com/science/article/pii/S0278612521001527},
author = {Tian Yu and Jing Huang and Qing Chang},
}

@misc{singh2023concurrentconstrainedoptimizationunknown,
      title={Concurrent Constrained Optimization of Unknown Rewards for Multi-Robot Task Allocation}, 
      author={Sukriti Singh and Anusha Srikanthan and Vivek Mallampati and Harish Ravichandar},
      year={2023},
      eprint={2305.15288},
      archivePrefix={arXiv},
      primaryClass={cs.RO},
      url={https://arxiv.org/abs/2305.15288}, 
}

@INPROCEEDINGS{7451735,
  author={Baraglia, Jimmy and Cakmak, Maya and Nagai, Yukie and Rao, Rajesh and Asada, Minoru},
  booktitle={HRI}, 
  title={Initiative in robot assistance during collaborative task execution}, 
  year={2016},
  volume={},
  number={},
  pages={67-74},
  doi={10.1109/HRI.2016.7451735}}

@InProceedings{Das2017ICCV,
author = {Das, Abhishek and Kottur, Satwik and Moura, Jose M. F. and Lee, Stefan and Batra, Dhruv},
title = {Learning Cooperative Visual Dialog Agents With Deep Reinforcement Learning},
booktitle = {Proceedings of the IEEE International Conference on Computer Vision (ICCV)},
month = {Oct},
year = {2017}
}

@misc{shervedani2024multimodalreinforcementlearningrobots,
      title={Multimodal Reinforcement Learning for Robots Collaborating with Humans}, 
      author={Afagh Mehri Shervedani and Siyu Li and Natawut Monaikul and Bahareh Abbasi and Barbara Di Eugenio and Milos Zefran},
      year={2024},
      eprint={2303.07265},
      archivePrefix={arXiv},
      primaryClass={cs.RO},
      url={https://arxiv.org/abs/2303.07265}, 
}

@inproceedings{padmakumar2022teach,
  title={Teach: Task-driven embodied agents that chat},
  author={Padmakumar, Aishwarya and Thomason, Jesse and Shrivastava, Ayush and Lange, Patrick and Narayan-Chen, Anjali and Gella, Spandana and Piramuthu, Robinson and Tur, Gokhan and Hakkani-Tur, Dilek},
  booktitle={Proceedings of the AAAI Conference on Artificial Intelligence},
  volume={36},
  number={2},
  pages={2017--2025},
  year={2022}
}

@article{Peters_2025,
   title={Emergent language: a survey and taxonomy},
   volume={39},
   ISSN={1573-7454},
   url={http://dx.doi.org/10.1007/s10458-025-09691-y},
   DOI={10.1007/s10458-025-09691-y},
   number={1},
   journal={Autonomous Agents and Multi-Agent Systems},
   publisher={Springer Science and Business Media LLC},
   author={Peters, Jannik and Waubert de Puiseau, Constantin and Tercan, Hasan and Gopikrishnan, Arya and Lucas de Carvalho, Gustavo Adolpho and Bitter, Christian and Meisen, Tobias},
   year={2025},
   month=mar }

@misc{lin2021learninggroundmultiagentcommunication,
      title={Learning to Ground Multi-Agent Communication with Autoencoders}, 
      author={Toru Lin and Minyoung Huh and Chris Stauffer and Ser-Nam Lim and Phillip Isola},
      year={2021},
      eprint={2110.15349},
      archivePrefix={arXiv},
      primaryClass={cs.LG},
      url={https://arxiv.org/abs/2110.15349}, 
}

@misc{lowe2020interactionsupervisionselfplayemergent,
      title={On the interaction between supervision and self-play in emergent communication}, 
      author={Ryan Lowe and Abhinav Gupta and Jakob Foerster and Douwe Kiela and Joelle Pineau},
      year={2020},
      eprint={2002.01093},
      archivePrefix={arXiv},
      primaryClass={cs.CL},
      url={https://arxiv.org/abs/2002.01093}, 
}

@inproceedings{kottur-etal-2017-natural,
    title = "Natural Language Does Not Emerge {\textquoteleft}Naturally' in Multi-Agent Dialog",
    author = "Kottur, Satwik  and
      Moura, Jos{\'e}  and
      Lee, Stefan  and
      Batra, Dhruv",
    editor = "Palmer, Martha  and
      Hwa, Rebecca  and
      Riedel, Sebastian",
    booktitle = "Proceedings of the 2017 Conference on Empirical Methods in Natural Language Processing",
    month = sep,
    year = "2017",
    address = "Copenhagen, Denmark",
    publisher = "Association for Computational Linguistics",
    url = "https://aclanthology.org/D17-1321/",
    doi = "10.18653/v1/D17-1321",
    pages = "2962--2967"
}

@misc{woodward2019learninginteractivelylearnassist,
      title={Learning to Interactively Learn and Assist}, 
      author={Mark Woodward and Chelsea Finn and Karol Hausman},
      year={2019},
      eprint={1906.10187},
      archivePrefix={arXiv},
      primaryClass={cs.AI},
      url={https://arxiv.org/abs/1906.10187}, 
}

@misc{lobostsunekawa2022madreamercoordinationcommunicationshared,
      title={MA-Dreamer: Coordination and communication through shared imagination}, 
      author={Kenzo Lobos-Tsunekawa and Akshay Srinivasan and Michael Spranger},
      year={2022},
      eprint={2204.04687},
      archivePrefix={arXiv},
      primaryClass={cs.LG},
      url={https://arxiv.org/abs/2204.04687}, 
}

@misc{kolb2019learningrequestguidanceemergent,
      title={Learning to Request Guidance in Emergent Communication}, 
      author={Benjamin Kolb and Leon Lang and Henning Bartsch and Arwin Gansekoele and Raymond Koopmanschap and Leonardo Romor and David Speck and Mathijs Mul and Elia Bruni},
      year={2019},
      eprint={1912.05525},
      archivePrefix={arXiv},
      primaryClass={cs.AI},
      url={https://arxiv.org/abs/1912.05525}, 
}

@inproceedings{codeaspolicies2022,
    title={Code as Policies: Language Model Programs for Embodied Control},
    author={Jacky Liang and Wenlong Huang and Fei Xia and Peng Xu and Karol Hausman and Brian Ichter and Pete Florence and Andy Zeng},
    booktitle={arXiv preprint arXiv:2209.07753},
    year={2022}
}

@misc{huang2023voxposercomposable3dvalue,
      title={VoxPoser: Composable 3D Value Maps for Robotic Manipulation with Language Models}, 
      author={Wenlong Huang and Chen Wang and Ruohan Zhang and Yunzhu Li and Jiajun Wu and Li Fei-Fei},
      year={2023},
      eprint={2307.05973},
      archivePrefix={arXiv},
      primaryClass={cs.RO},
      url={https://arxiv.org/abs/2307.05973}, 
}

@inproceedings{luo2023obtaining,
  title={Obtaining hierarchy from human instructions: an llms-based approach},
  author={Luo, Xusheng and Xu, Shaojun and Liu, Changliu},
  booktitle={CoRL 2023 Workshop on Learning Effective Abstractions for Planning (LEAP)},
  year={2023}
}

@article{choi2025reactree,
  title={Reactree: Hierarchical task planning with dynamic tree expansion using llm agent nodes},
  author={Choi, Jae-Woo and Kim, Hyungmin and Ong, Hyobin and Yoon, Youngwoo and Jang, Minsu and Kim, Jaehong and others},
  year={2025}
}

@InProceedings{pmlrv235li24ar,
  title = 	 {Chain of Code: Reasoning with a Language Model-Augmented Code Emulator},
  author =       {Li, Chengshu and Liang, Jacky and Zeng, Andy and Chen, Xinyun and Hausman, Karol and Sadigh, Dorsa and Levine, Sergey and Fei-Fei, Li and Xia, Fei and Ichter, Brian},
  booktitle = 	 {Proceedings of the 41st International Conference on Machine Learning},
  pages = 	 {28259--28277},
  year = 	 {2024},
  editor = 	 {Salakhutdinov, Ruslan and Kolter, Zico and Heller, Katherine and Weller, Adrian and Oliver, Nuria and Scarlett, Jonathan and Berkenkamp, Felix},
  volume = 	 {235},
  series = 	 {Proceedings of Machine Learning Research},
  month = 	 {21--27 Jul},
  publisher =    {PMLR},
  url = 	 {https://proceedings.mlr.press/v235/li24ar.html},
}

@misc{belkhale2024rthactionhierarchiesusing,
      title={RT-H: Action Hierarchies Using Language}, 
      author={Suneel Belkhale and Tianli Ding and Ted Xiao and Pierre Sermanet and Quon Vuong and Jonathan Tompson and Yevgen Chebotar and Debidatta Dwibedi and Dorsa Sadigh},
      year={2024},
      eprint={2403.01823},
      archivePrefix={arXiv},
      primaryClass={cs.RO},
      url={https://arxiv.org/abs/2403.01823}, 
}

@misc{shi2024yellrobotimprovingonthefly,
      title={Yell At Your Robot: Improving On-the-Fly from Language Corrections}, 
      author={Lucy Xiaoyang Shi and Zheyuan Hu and Tony Z. Zhao and Archit Sharma and Karl Pertsch and Jianlan Luo and Sergey Levine and Chelsea Finn},
      year={2024},
      eprint={2403.12910},
      archivePrefix={arXiv},
      primaryClass={cs.RO},
      url={https://arxiv.org/abs/2403.12910}, 
}

@misc{shi2025hirobotopenendedinstruction,
      title={Hi Robot: Open-Ended Instruction Following with Hierarchical Vision-Language-Action Models}, 
      author={Lucy Xiaoyang Shi and Brian Ichter and Michael Equi and Liyiming Ke and Karl Pertsch and Quan Vuong and James Tanner and Anna Walling and Haohuan Wang and Niccolo Fusai and Adrian Li-Bell and Danny Driess and Lachy Groom and Sergey Levine and Chelsea Finn},
      year={2025},
      eprint={2502.19417},
      archivePrefix={arXiv},
      primaryClass={cs.RO},
      url={https://arxiv.org/abs/2502.19417}, 
}

@misc{ma2024eurekahumanlevelrewarddesign,
      title={Eureka: Human-Level Reward Design via Coding Large Language Models}, 
      author={Yecheng Jason Ma and William Liang and Guanzhi Wang and De-An Huang and Osbert Bastani and Dinesh Jayaraman and Yuke Zhu and Linxi Fan and Anima Anandkumar},
      year={2024},
      eprint={2310.12931},
      archivePrefix={arXiv},
      primaryClass={cs.RO},
      url={https://arxiv.org/abs/2310.12931}, 
}

@inproceedings{ma2024dreureka,
    title   = {DrEureka: Language Model Guided Sim-To-Real Transfer},
    author  = {Yecheng Jason Ma and William Liang and Hungju Wang and Sam Wang and Yuke Zhu and Linxi Fan and Osbert Bastani and Dinesh Jayaraman},
    year    = {2024},
  booktitle = {Robotics: Science and Systems (RSS)}
}

@misc{yu2023languagerewardsroboticskill,
      title={Language to Rewards for Robotic Skill Synthesis}, 
      author={Wenhao Yu and Nimrod Gileadi and Chuyuan Fu and Sean Kirmani and Kuang-Huei Lee and Montse Gonzalez Arenas and Hao-Tien Lewis Chiang and Tom Erez and Leonard Hasenclever and Jan Humplik and Brian Ichter and Ted Xiao and Peng Xu and Andy Zeng and Tingnan Zhang and Nicolas Heess and Dorsa Sadigh and Jie Tan and Yuval Tassa and Fei Xia},
      year={2023},
      eprint={2306.08647},
      archivePrefix={arXiv},
      primaryClass={cs.RO},
      url={https://arxiv.org/abs/2306.08647}, 
}

@INPROCEEDINGS{6630673,

  author={Knepper, Ross A. and Layton, Todd and Romanishin, John and Rus, Daniela},

  booktitle={2013 IEEE International Conference on Robotics and Automation}, 

  title={IkeaBot: An autonomous multi-robot coordinated furniture assembly system}, 

  year={2013},

  volume={},

  number={},

  pages={855-862},

  doi={10.1109/ICRA.2013.6630673}}

@inproceedings{10.5555/2832747.2832901,
author = {Veloso, Manuela and Biswas, Joydeep and Coltin, Brian and Rosenthal, Stephanie},
title = {CoBots: robust symbiotic autonomous mobile service robots},
year = {2015},
isbn = {9781577357384},
publisher = {AAAI Press},
booktitle = {IJCAI, 2015},
pages = {4423–4429},
numpages = {7},
location = {Buenos Aires, Argentina},
series = {}
}

@article{chen:jair10,
title={Training a Multilingual Sportscaster: Using Perceptual Context to Learn Language},
author={David L. Chen and Joohyun Kim and Raymond J. Mooney},
volume={37},
journal={Journal of Artificial Intelligence Research},
pages={397--435},
url="http://www.cs.utexas.edu/users/ai-lab?chen:jair10",
year={2010}
}

@INPROCEEDINGS{4115652,

  author={Mutlu, Bilge and Forlizzi, Jodi and Hodgins, Jessica},

  booktitle={2006 6th IEEE-RAS International Conference on Humanoid Robots}, 

  title={A Storytelling Robot: Modeling and Evaluation of Human-like Gaze Behavior}, 

  year={2006},

  volume={},

  number={},

  pages={518-523},

  doi={10.1109/ICHR.2006.321322}}

@article{cascianelli2018full,
  title={Full-GRU natural language video description for service robotics applications},
  author={Cascianelli, Silvia and Costante, Gabriele and Ciarfuglia, Thomas A and Valigi, Paolo and Fravolini, Mario L},
  journal={IEEE RA-L},
  volume={3},
  number={2},
  pages={841--848},
  year={2018},
  publisher={IEEE}
}

@inproceedings{
li2022behavior,
title={{BEHAVIOR}-1K: A Benchmark for Embodied {AI} with 1,000 Everyday Activities and Realistic Simulation},
author={Chengshu Li and Ruohan Zhang and Josiah Wong and Cem Gokmen and Sanjana Srivastava and Roberto Mart{\'\i}n-Mart{\'\i}n and Chen Wang and Gabrael Levine and Michael Lingelbach and Jiankai Sun and Mona Anvari and Minjune Hwang and Manasi Sharma and Arman Aydin and Dhruva Bansal and Samuel Hunter and Kyu-Young Kim and Alan Lou and Caleb R Matthews and Ivan Villa-Renteria and Jerry Huayang Tang and Claire Tang and Fei Xia and Silvio Savarese and Hyowon Gweon and Karen Liu and Jiajun Wu and Li Fei-Fei},
booktitle={CoRL},
year={2022},
url={https://openreview.net/forum?id=_8DoIe8G3t}
}

@misc{myers2025learningtemporaldistancescontrastive,
      title={Learning Temporal Distances: Contrastive Successor Features Can Provide a Metric Structure for Decision-Making}, 
      author={Vivek Myers and Chongyi Zheng and Anca Dragan and Sergey Levine and Benjamin Eysenbach},
      year={2025},
      eprint={2406.17098},
      archivePrefix={arXiv},
      primaryClass={cs.LG},
      url={https://arxiv.org/abs/2406.17098}, 
}

@article{shah2024bumble,
   title={BUMBLE: Unifying Reasoning and Acting with Vision-Language Models for Building-wide Mobile Manipulation},
   author={Shah, Rutav and Yu, Albert and Zhu, Yifeng and Zhu, Yuke and Mart{\'\i}n-Mart{\'\i}n, Roberto},
   journal={ICRA},
   year={2025},
   organization={IEEE}
}

@article{jin2023minibehavior,
      title={Mini-BEHAVIOR: A Procedurally Generated Benchmark for Long-horizon Decision-Making in Embodied AI}, 
      author={Emily Jin and Jiaheng Hu and Zhuoyi Huang and Ruohan Zhang and Jiajun Wu and Li Fei-Fei and Roberto Mart{\'i}n-Mart{\'i}n},
      year={2023},
      journal={arXiv preprint 2310.01824},
}

@article{ouyang2022training,
  title={Training language models to follow instructions with human feedback},
  author={Ouyang, Long and Wu, Jeffrey and Jiang, Xu and Almeida, Diogo and Wainwright, Carroll and Mishkin, Pamela and Zhang, Chong and Agarwal, Sandhini and Slama, Katarina and Ray, Alex and others},
  journal={NeurIPS},
  volume={35},
  pages={27730--27744},
  year={2022}
}

@article{achiam2023gpt,
  title={Gpt-4 technical report},
  author={Achiam, Josh and Adler, Steven and Agarwal, Sandhini and Ahmad, Lama and Akkaya, Ilge and Aleman, Florencia Leoni and Almeida, Diogo and Altenschmidt, Janko and Altman, Sam and Anadkat, Shyamal and others},
  journal={arXiv preprint arXiv:2303.08774},
  year={2023}
}

@article{selvaggio2021autonomy,
  title={Autonomy in physical human-robot interaction: A brief survey},
  author={Selvaggio, Mario and Cognetti, Marco and Nikolaidis, Stefanos and Ivaldi, Serena and Siciliano, Bruno},
  journal={IEEE RA-L},
  volume={6},
  number={4},
  pages={7989--7996},
  year={2021},
  publisher={IEEE}
}

@article{deng2023survey,
  title={A survey on proactive dialogue systems: Problems, methods, and prospects},
  author={Deng, Yang and Lei, Wenqiang and Lam, Wai and Chua, Tat-Seng},
  journal={arXiv preprint arXiv:2305.02750},
  year={2023}
}

@article{chen2023controllable,
  title={Controllable mixed-initiative dialogue generation through prompting},
  author={Chen, Maximillian and Yu, Xiao and Shi, Weiyan and Awasthi, Urvi and Yu, Zhou},
  journal={arXiv preprint arXiv:2305.04147},
  year={2023}
}

@inproceedings{yu2024lang4sim2real,
      title={Natural Language Can Help Bridge the Sim2Real Gap},
      author={Yu, Albert and Foote, Adeline and Mooney, Raymond and Martín-Martín, Roberto},
      booktitle={Robotics: Science and Systems (RSS), 2024},
      year={2024}
}

@inproceedings{collabllm2025,
    title={CollabLLM: From Passive Responders to Active Collaborators},
    author={Shirley Wu and Michel Galley and Baolin Peng and Hao Cheng and 
            Gavin Li and Yao Dou and Weixin Cai and James Zou and 
            Jure Leskovec and Jianfeng Gao},
    booktitle={ICML},
    year={2025}
}

@article{Thomason2019ImprovingGN,
  title={Improving Grounded Natural Language Understanding through Human-Robot Dialog},
  author={Jesse Thomason and Aishwarya Padmakumar and Jivko Sinapov and Nick Walker and Yuqian Jiang and Harel Yedidsion and Justin W. Hart and Peter Stone and Raymond J. Mooney},
  journal={2019 International Conference on Robotics and Automation (ICRA)},
  year={2019},
  pages={6934-6941},
  url={https://api.semanticscholar.org/CorpusID:67703525}
}


\end{document}